# Compiling Uncertainty Away in Conformant Planning Problems with Bounded Width

**Hector Palacios**                                                    HLP@LDC.USB.VE
*Universitat Pompeu Fabra*
*Roc Boronat, 138*
*08018 Barcelona, SPAIN*

**Hector Geffner**                                          HECTOR.GEFFNER@UPF.EDU
*ICREA & Universitat Pompeu Fabra*
*Roc Boronat, 138*
*08018 Barcelona, SPAIN*

## Abstract

Conformant planning is the problem of finding a sequence of actions for achieving a goal in the presence of uncertainty in the initial state or action effects. The problem has been approached as a path-finding problem in belief space where good belief representations and heuristics are critical for scaling up. In this work, a different formulation is introduced for conformant problems with deterministic actions where they are automatically converted into classical ones and solved by an off-the-shelf classical planner. The translation maps literals $L$ and sets of assumptions $t$ about the initial situation, into new literals $KL/t$ that represent that $L$ must be true if $t$ is initially true. We lay out a general translation scheme that is sound and establish the conditions under which the translation is also complete. We show that the complexity of the complete translation is exponential in a parameter of the problem called the *conformant width*, which for most benchmarks is bounded. The planner based on this translation exhibits good performance in comparison with existing planners, and is the basis for $T_0$, the best performing planner in the Conformant Track of the 2006 International Planning Competition.

## 1. Introduction

Conformant planning is a form of planning where a goal is to be achieved when the initial situation is not fully known and actions may have non-deterministic effects (Goldman & Boddy, 1996; Smith & Weld, 1998). Conformant planning is computationally harder than classical planning, as even under polynomial restrictions on plan length, plan verification remains hard (Haslum & Jonsson, 1999; Baral, Kreinovich, & Trejo, 2000; Turner, 2002; Rintanen, 2004). While few practical problems are purely conformant, the ability to find conformant plans is needed in contingent planning where conformant situations are a special case and where relaxations into conformant planning yield useful heuristics (Hoffmann & Brafman, 2005).

The problem of conformant planning can be formulated as a path-finding problem in belief space where a sequence of actions that map a given initial belief state into a target belief is sought (Bonet & Geffner, 2000). A belief state represents the set of states that are deemed possible, and actions, whether deterministic or not, map one belief state into another. This formulation, that underlies most current conformant planners (Hoffmann & Brafman, 2006; Bryce, Kambhampati, & Smith, 2006; Cimatti, Roveri, & Bertoli, 2004) must address two problems: the problem of representing beliefs in a compact way, and the problem of obtaining effective heuristics over beliefs. The first problem has been approached through logical representations that make use of SAT or OBDD technology, that while intractable in the worst case, scale up better than plain state representations. The second problem, on the other hand, has been more complex, with heuristics for searching in belief space not being as successful so far as the heuristics developed for classical planning (Bonet & Geffner, 2001; Hoffmann & Nebel, 2001).

In this work, we introduce a different approach to conformant planning where problems are automatically compiled into classical problems and solved by a classical planner. The translation maps sets of literals $t$ about the initial situation and literals $L$ into new literals $KL/t$ that express that if $t$ is true in the initial situation, $L$ must be true. We lay out first a general translation scheme that is sound and then establish the conditions under which the translation is also complete. Also, we show that the complexity of the complete translation is exponential in a parameter of the problem that we call the *conformant width*, which for most benchmark domains is bounded, implying that the complete translation in those cases is polynomial. The planner based on this translation exhibits good performance in comparison with existing conformant planners and is the basis for $T_0$, the best performing planner in the Conformant Track of the 2006 International Planning Competition.

The translation-based approach provides a solution to the two problems faced by conformant planners that search in belief space: the belief representation and the heuristic over beliefs. In the translation-based approach, the beliefs are represented by the literals $KL/t$ that stand for conditionals, a representation that is polynomial and complete for conformant problems with bounded width. In addition, and since belief states are represented as plain states, the heuristic over beliefs is a classical heuristic. From a computational point of view, though, there is no explicit search in belief-space: conformant problems $P$ are converted into classical problems $K(P)$ at the 'knowledge-level' (Petrick & Bacchus, 2002), whose solutions, computed by a classical planner, encode the conformant solutions for $P$.

Our formulation is limited to conformant problems that are *deterministic* and where all uncertainty lies in the initial situation. We address nonetheless the issues that must be handled in order to generalize the translation-based approach to non-deterministic domains and report empirical results over non-deterministic domains as well.

The paper is organized as follows. We define first the syntax and semantics of conformant planning problems $P$ (Section 2), and consider a simple sound but incomplete translation $K_0$ (Section 3). We then consider a more general translation scheme $K_{T,M}$ where $T$ and $M$ are two parameters, a set of tags $t$ encoding assumptions about the initial situation, and a set of merges $m$ encoding valid disjunctions of tags (Section 4), and analyze several instances of this scheme that follow from particular choices of the sets of tags and merges: a complete but exponential translation $K_{S0}$ where tags are associated with the possible initial states of the problem (Section 5), and a polynomial translation $K_i$ for a fixed integer $i \geq 0$ that is complete for problems with conformant width bounded by $i$ (Section 6). We provide then an alternative explanation for this compact but complete translation by showing that in problems with bounded width, the exponential number of possible initial states $S_0$ includes always a polynomial number of 'critical' initial states $S_0'$ such that plans that conform with $S_0'$ conform also with $S_0$ (Section 7). We finally present the conformant planner $T_0$ (Section 8), an empirical evaluation of the planner (Section 9), an extension to non-deterministic actions (Section 10), and a discussion of related work (Section 11). This is followed by a brief summary (Section 12) and the formal proofs (Appendix).

This work is a revision and extension of the formulation presented by Palacios and Geffner (2007), which in turn is based on ideas first sketched also by Palacios and Geffner (2006).

## 2. The Conformant Problem $P$

We define next the syntax and semantics of the conformant planning problems considered.

### 2.1 Syntax

Conformant planning problems $P$ are represented as tuples of the form $P = \langle F, I, O, G \rangle$ where $F$ stands for the fluent symbols in the problem, $I$ is a set of clauses over $F$ defining the initial situation, $O$ stands for a set of (ground) operators or actions $a$, and $G$ is a set of literals over $F$ defining the goal. Every action $a$ has a precondition $Pre(a)$ given by a set of fluent literals, and a set of conditional effects $C \rightarrow L$ where $C$ is a set of fluent literals and $L$ is a fluent literal.

All actions are assumed to be *deterministic* and hence all uncertainty lies in the initial situation. Thus, the language for the conformant problem $P$ excluding the uncertainty in the initial situation, is Strips extended with conditional effects and negation. Moreover, if there is no uncertainty in the initial situation, as when all fluents appear in unit clauses in $I$, $P$ is equivalent to a classical planning problem.

We refer to the conditional effects $C \rightarrow L$ of an action $a$ as the *rules* associated with $a$, and sometimes write them as $a : C \rightarrow L$. When convenient, we also join several effects associated with the same action and condition as in $a : C \rightarrow L \wedge L'$ and write $C \rightarrow L$ as $true \rightarrow L$ when $C$ is empty. Finally, for a literal $L$, $\neg L$ denotes the complement of $L$.

### 2.2 Semantics

A state $s$ is a truth assignment over the fluents $F$ in $P = \langle F, I, O, G \rangle$ and a possible initial state $s$ of $P$ is a state that satisfies the clauses in $I$.

For a state $s$, we write $I(s)$ to refer to the set of atoms (positive literals) that are true in $s$, and write $P/s$ to refer to the *classical planning problem* $P/s = \langle F, I(s), O, G \rangle$ which is like the conformant problem $P$ except for the initial state that is fixed to $s$.

An action sequence $\pi = \{a_0, a_1, \ldots, a_n\}$ is a *classical plan* for $P/s$ if the action sequence $\pi$ is executable in the state $s$ and results in a goal state $s_G$; i.e., if for $i = 0, \ldots, n$, the preconditions of the action $a_i$ are true in $s_i$, $s_{i+1}$ is the state that results from doing action $a_i$ in the state $s_i$, and all goal literals are true in $s_{n+1}$.

Finally, an action sequence $\pi$ is a *conformant plan* for $P$ iff $\pi$ is a classical plan for $P/s$ for every possible initial state $s$ of $P$.

Conformant planning is computationally harder than classical planning, as plan verification remains hard even under polynomial restrictions on plan length (Haslum & Jonsson, 1999; Baral et al., 2000; Turner, 2002; Rintanen, 2004). The most common approach to conformant planning is based on the belief state formulation (Bonet & Geffner, 2000). A belief state $b$ is the non-empty set of states that are deemed possible in a given situation, and every action $a$ executable in $b$, maps $b$ into a new belief state $b_a$. The conformant planning task becomes a path-finding problem in a graph where the nodes are belief states $b$, the source node $b_0$ is the belief state corresponding to the initial situation, and the target belief states $b_G$ are those where the goals are true.

We assume throughout that $I$ is logically consistent, so that the set of possible initial states is not empty, and that $P$ itself is *consistent*, so that the bodies $C$ and $C'$ of conflicting effects $a : C \rightarrow L$ and $a : C' \rightarrow \neg L$ associated with the same action $a$ are mutually exclusive or mutex. For further details on this; see Part B of the Appendix.

## 3. A Basic Translation $K_0$

A simple translation of the conformant problem $P$ into a classical problem $K(P)$ can be obtained by replacing the literals $L$ by literals $KL$ and $K\neg L$ aimed at capturing whether $L$ is known to be true and known to be false respectively.

**Definition 1** (Translation $K_0$). For a conformant planning problem $P = \langle F, I, O, G \rangle$, the translation $K_0(P) = \langle F', I', O', G' \rangle$ is a classical planning problem with

- $F' = \{KL, K\neg L \mid L \in F\}$
- $I' = \{KL \mid L \text{ is a unit clause in } I\}$
- $G' = \{KL \mid L \in G\}$
- $O' = O$ but with each precondition $L$ for $a \in O$ replaced by $KL$, and each conditional effect $a : C \rightarrow L$ replaced by $a : KC \rightarrow KL$ and $a : \neg K\neg C \rightarrow \neg K\neg L$, where the expressions $KC$ and $\neg K\neg C$ for $C = L_1, L_2 \ldots$ are abbreviations of the formulas $KL_1, KL_2 \ldots$ and $\neg K\neg L_1, \neg K\neg L_2 \ldots$ respectively.

The intuition behind the translation is simple: first, the literal $KL$ is true in the initial state $I'$ if $L$ is known to be true in $I$; otherwise it is false. This removes all uncertainty from $K_0(P)$, making it into a classical planning problem. In addition, for soundness, each rule $a : C \rightarrow L$ in $P$ is mapped into *two* rules: a **support rule** $a : KC \rightarrow KL$, that ensures that $L$ is known to be true when the condition is known to be true, and a **cancellation rule** $a : \neg K\neg C \rightarrow \neg K\neg L$ that guarantees that $K\neg L$ is deleted (prevented to persist) when action $a$ is applied and $C$ is not known to be false. The use of support and cancellation rules for encoding the original rules at the 'knowledge-level' is the only subtlety in the translation.

The translation $K_0(P)$ is sound as every classical plan that solves $K_0(P)$ is a conformant plan for $P$, but is incomplete, as not all conformant plans for $P$ are classical plans for $K(P)$. The meaning of the $KL$ literals follows a similar pattern: if a plan achieves $KL$ in $K_0(P)$, then the same plan achieves $L$ with certainty in $P$, yet a plan may achieve $L$ with certainty in $P$ without making the literal $KL$ true in $K_0(P)$.[1]

**Proposition 2** (Soundness of $K_0(P)$). *If $\pi$ is a classical plan for $K_0(P)$, then $\pi$ is a conformant plan for $P$.*

---

[1]. Formal proofs can be found in the appendix.

As an illustration, consider the conformant problem $P = \langle F, I, O, G \rangle$ with $F = \{p, q, r\}$, $I = \{q\}$, $G = \{p, r\}$, and actions $O = \{a, b\}$ with effects

$$a : q \rightarrow r \ , \ a : p \rightarrow \neg p \ , \ b : q \rightarrow p \ .$$

For this problem, the action sequence $\pi = \{a, b\}$ is a conformant plan for $P$ while the action sequence $\pi' = \{a\}$ is not. Indeed, $\pi$ is a classical plan for $P/s$ for any possible initial state $s$, while $\pi'$ is not a classical plan for the possible initial state $s'$ where $p$ is true (recall that $s$ is a possible initial state of $P$ if $s$ satisfies $I$ so that neither $p$ nor $r$ are assumed to be initially false in this problem).

From Definition 1, the translation $K_0(P) = \langle F', I', O', G' \rangle$ is a classical planning problem with fluents $F' = \{Kp, K\neg p, Kq, K\neg q, Kr, K\neg r\}$, initial situation $I' = \{Kq\}$, goals $G' = \{Kp, Kr\}$, and actions $O' = \{a, b\}$ with effects

$$a : Kq \rightarrow Kr \ , \ a : Kp \rightarrow K\neg p \ , \ b : Kq \rightarrow Kp,$$

that encode supports, and effects

$$a : \neg K\neg q \rightarrow \neg K\neg r \ , \ a : \neg K\neg p \rightarrow \neg Kp \ , \ b : \neg K\neg q \rightarrow \neg K\neg p,$$

that encode cancellations.

Proposition 2 implies, for example, that $\pi' = \{a\}$, which is not a conformant plan for $P$, cannot be a classical plan for $K(P)$ either. This is easy to verify, as while the support $a : Kq \rightarrow Kr$ achieves the goal $Kr$ as $Kq$ is true in $I'$, the cancellation $a : \neg K\neg p \rightarrow \neg Kp$ associated with the same action, preserves $Kp$ false for the other goal $p$.

While the translation $K_0$ is not complete, meaning that it fails to capture all conformant plans for $P$ as classical plans, its completeness can be assessed in terms of a weaker semantics. In the so-called 0-approximation semantics (Baral & Son, 1997), belief states $b$ are represented by 3-valued states where fluents can be true, false, or unknown. In this incomplete belief representation, checking whether an action $a$ is applicable in a belief state $b$, computing the next belief state $b_a$, and verifying polynomial length plans are all polynomial time operations. In particular, a literal $L$ is true it the next belief state $b_a$ iff a) action $a$ has some effect $C \rightarrow L$ such that all literals in $C$ are true in $b$, or b) $L$ is true in $b$ and for all effects $C' \rightarrow \neg L$ of action $a$, the complement of some literal $L' \in C'$ is true in $b$. An action sequence $\pi$ is then *a conformant plan for $P$ according to the 0-approximation semantics* if the belief sequence generated by $\pi$ according to the 0-approximation semantics makes the action sequence applicable and terminates in a belief state where the goals are true. It is possible to prove then that:

**Proposition 3** ($K_0(P)$ and 0-Approximation). *An action sequence $\pi$ is a classical plan for $K_0(P)$ iff $\pi$ is a conformant plan for $P$ according to the 0-approximation semantics.*

This correspondence is not surprising though as both the 0-approximation semantics and the $K_0(P)$ translation throw away the disjunctive information and restrict the plans to those that make no use of the uncertain knowledge. Indeed, the states $s_0, s_1, \ldots$ generated by the action sequence $\pi = \{a_0, a_1, \ldots\}$ over the classical problem $K_0(P)$ encode precisely the literals that are known to be true according to the 0-approximation; namely, $L$ is true at time $i$ according to the 0-approximation iff the literal $KL$ is true in the state $s_i$.

Proposition 3 does not mean that the basic translation $K_0$ and the 0-approximation semantics are equivalent but rather that they both rely on equivalent belief representations. The translation $K_0$ delivers also a way to get valid conformant plans using a classical planner. The translation-based approach thus addresses both the representational and the heuristic issues that arise in conformant planning.

As an illustration of Proposition 3, given a conformant problem $P$ with $I = \{p, r\}$ and actions $a$ and $b$ with effects $a : p \to q$, $a : r \to \neg v$, and $b : q \to v$, the plan $\pi = \{a, b\}$ is valid for achieving the goal $G = \{q, v\}$ according to both $K_0(P)$ and the 0-approximation, while the plan $\pi = \{b\}$ is not valid according to either. At the same time, if the initial situation is changed to $I = \{p \vee q\}$, neither approach sanctions the plan $\pi = \{a\}$ for $G = \{q\}$, even if it is a valid conformant plan. For this, some ability to reason with disjunctions is needed.

An extension of the basic translation $K_0$ that allows a limited form of disjunctive reasoning is presented by Palacios and Geffner (2006). The extension is based on the introduction of new literals $L/X_i$ used for encoding the conditionals $X_i \supset L$. Below, the basic translation $K_0$ is extended in a different manner that ensures both tractability and completeness over a large class of problems.

## 4. General Translation Scheme $K_{T,M}$

The basic translation $K_0$ is extended now into a general translation scheme $K_{T,M}$ where $T$ and $M$ are two parameters: a set of *tags* $t$ and a set of merges $m$. We will show that for suitable choices of these two parameters, the translation $K_{T,M}$, unlike the translation $K_0$, can be both sound and complete.

A tag $t \in T$ is a set (conjunction) of literals $L$ from $P$ whose truth value in the initial situation is not known. The tags $t$ are used to introduce a new class of literals $KL/t$ in the classical problem $K_{T,M}(P)$ that represent the conditional 'if $t$ is true initially, then $L$ is true', an assertion that could be written as $K(t_0 \supset L)$ in a temporal modal logic. We use the notation $KL/t$ rather than $L/t$ as used by Palacios and Geffner (2006), because there is a distinction between $\neg KL/t$ and $K\neg L/t$: roughly $\neg KL/t$ means that the conditional $K(t_0 \supset L)$ is not true, while $K\neg L/t$ means that the conditional $K(t_0 \supset \neg L)$ is true.

Likewise, a merge $m$ is a non-empty collection of tags $t$ in $T$ that stands for the Disjunctive Normal Form (DNF) formula $\bigvee_{t \in m} t$. A merge $m$ is *valid* when one of the tags $t \in m$ must be true in $I$; i.e., when

$$I \models \bigvee_{t \in m} t \quad .$$

A merge $m$ for a literal $L$ in $P$ will translate into a 'merge action' with a single effect

$$\bigwedge_{t \in m} KL/t \;\; \to \;\; KL$$

that captures a simple form of reasoning by cases.

While a valid merge can be used for reasoning about any literal $L$ in $P$, computationally it is convenient (although not logically necessary) to specify that certain merges are to be used with some literals $L$ and not with others. Thus, formally, $M$ is a collection of pairs $(m, L)$, where $m$ is a merge and $L$ is a literal in $P$. Such a pair means that $m$ is a merge for $L$. We group all the merges $m$ for a literal $L$ in the set $M_L$, and thus, $M$ can be understood as the collection of such sets $M_L$ for all $L$ in $P$. For simplicity, however, except when it may cause a confusion, we will keep referring to $M$ as a plain set of merges.

We assume that the collection of tags $T$ always includes a tag $t$ that stands for the empty collection of literals, that we call the *empty tag* and denote it as $\emptyset$. If $t$ is the empty tag, we denote $KL/t$ simply as $KL$.

The translation $K_{T,M}(P)$ is the basic translation $K_0(P)$ 'conditioned' with the tags $t$ in $T$ and extended with the actions that capture the merges in $M$:

**Definition 4** (Translation $K_{T,M}$). Let $P = \langle F, I, O, G \rangle$ be a conformant problem, then $K_{T,M}(P)$ is the *classical planning problem* $K_{T,M}(P) = \langle F', I', O', G' \rangle$ with

- $F' = \{KL/t, K\neg L/t \mid L \in F \text{ and } t \in T\}$

- $I' = \{KL/t \mid I, t \models L\}$

- $G' = \{KL \mid L \in G\}$

- $O' = \{a : KC/t \to KL/t,\ a : \neg K\neg C/t \to \neg K\neg L/t \mid a : C \to L \text{ in } P\}\ \cup$
  $\{a_{m,L} : [\bigwedge_{t \in m} KL/t] \to KL \wedge XL \mid L \in P, m \in M_L\}$

where $KL$ is a precondition of action $a$ in $K_{T,M}(P)$ if $L$ is a precondition of $a$ in $P$, $KC/t$ and $\neg K\neg C/t$ stand for $KL_1/t, KL_2/t, \ldots$, and $\neg K\neg L_1/t, \neg K\neg L_2/t, \ldots$ respectively, when $C = L_1, L_2, \ldots$, and $XL$ stands for $\bigwedge_{L'} K\neg L'$ with $L'$ ranging over the literals $L'$ mutex with $L$ in $P$.

The translation $K_{T,M}(P)$ reduces to the basic translation $K_0(P)$ when $M$ is empty and $T$ contains only the empty tag. The extra effects $XL = \bigwedge_{L'} K\neg L'$ in the merge actions $a_{m,L}$ are needed only to ensure that the translation $K_{T,M}(P)$ is consistent when $P$ is consistent, and otherwise can be ignored. Indeed, if $L$ and $L'$ are mutex in a consistent $P$, the invariant $KL/t \supset K\neg L'/t$ holds in $K_{T,M}(P)$ for non-empty tags $t$, and hence a successful merge for $L$ can always be followed by a successful merge for $\neg L'$. In the rest of the paper we will thus assume that both $P$ and $K_{T,M}(P)$ are consistent, and ignore such extra merge effects, but we will come back to them in Appendix B for proving the consistency of $K_{T,M}(P)$ from the consistency of $P$.

For suitable choices of $T$ and $M$, the translation $K_{T,M}(P)$ will be *sound and complete*. Before establishing these results, however, let us make these notions precise.

**Definition 5** (Soundness). A translation $K_{T,M}(P)$ is sound if for any classical plan $\pi$ that solves the classical planning problem $K_{T,M}(P)$, the plan $\pi'$ that results from $\pi$ by dropping the merge actions is a conformant plan for $P$.

**Definition 6** (Completeness). A translation $K_{T,M}(P)$ is complete if for any conformant plan $\pi'$ that solves the conformant problem $P$, there is a classical plan $\pi$ that solves the classical problem $K_{T,M}(P)$ such that $\pi'$ is equal to $\pi$ with the merge actions removed.

The general translation scheme $K_{T,M}$ is sound provided that all merges are valid and all tags are consistent (literals in a tag are all true in some possible initial state):

**Theorem 7** (Soundness $K_{T,M}(P)$)**.** *The translation $K_{T,M}(P)$ is sound provided that all merges in $M$ are valid and all tags in $T$ are consistent.*

Unless stated otherwise, we will assume that all merges are valid and all tags consistent, and will call such translations, *valid translations*.

As a convention for keeping the notation simple, in singleton tags like $t = \{p\}$, the curly brackets are often dropped. Thus, literals $KL/t$ for $t = \{p\}$ are written as $KL/p$, while merges $m = \{t_1, t_2\}$ for singleton tags $t_1 = \{p\}$ and $t_2 = \{q\}$, are written as $m = \{p, q\}$.

**Example.** As an illustration, consider the problem of moving an object from an origin to a destination using two actions: $pick(l)$, that picks up an object from a location if the hand is empty and the object is in that location, and $drop(l)$, that drops the object at a location if the object is being held. For making the problem more interesting, let us also assume that the action $pick(l)$ drops the object being held at $l$ if the hand is not empty. These are all conditional effects and there are no action preconditions. Assuming that there is a single object, these effects can be written as:

$$pick(l) : \neg hold, at(l) \rightarrow hold \wedge \neg at(l)$$
$$pick(l) : hold \rightarrow \neg hold \wedge at(l)$$
$$drop(l) : hold \rightarrow \neg hold \wedge at(l) \ .$$

Consider now an instance $P$ of this domain, where the hand is initially empty and the object, initially at either $l_1$ or $l_2$, must be moved to $l_3$; i.e., $P = \langle F, I, O, G \rangle$ with

$$I = \{\neg hold, at(l_1) \vee at(l_2), \neg at(l_1) \vee \neg at(l_2), \neg at(l_3)\}$$

and

$$G = \{at(l_3)\} \ .$$

The action sequence

$$\pi_1 = \{pick(l_1), drop(l_3), pick(l_2), drop(l_3)\}$$

is a conformant plan for this problem, where an attempt to pick up the object at location $l_1$ is followed by a drop at the target location $l_3$, ensuring that the object ends up at $l_3$ if it was originally at $l_1$. This is then followed by an attempt to pick up the object at $l_2$ and a drop at $l_3$.

On the other hand, the action sequence $\pi_2$ that results from $\pi_1$ by removing the first drop action

$$\pi_2 = \{pick(l_1), pick(l_2), drop(l_3)\}$$

is not a conformant plan, since if the object was originally at $l_1$, it would end up at $l_2$ after the action $pick(l_2)$. In the notation introduced above, $\pi_1$ is a classical plan for the classical problem $P/s$ for the two possible initial states $s$, while $\pi_2$ is a classical plan for the problem $P/s$ but only for the state $s$ where the object is initially at $l_2$.

Consider now the classical problem $K_{T,M}(P) = \langle F', I', O', G' \rangle$ that is obtained from $P$ when $T = \{at(l_1), at(l_2)\}^2$ and $M$ contains the merge $m = \{at(l_1), at(l_2)\}$ for the literals *hold* and $at(l_3)$. From its definition, the fluents $F'$ in $K_{T,M}(P)$ are of the form $KL/t$ and $K\neg L/t$ for $L \in \{at(l), hold\}$, $l \in \{l_1, l_2\}$, and $t \in T$, while the initial situation $I'$ is

$$I' = \{K\neg hold, K\neg hold/at(l), K\neg at(l_3), K\neg at(l_3)/at(l), Kat(l)/at(l), K\neg at(l')/at(l)\}$$

for $l, l' \in \{l_1, l_2\}$ and $l' \neq l$, and the goal $G'$ is

$$G' = \{Kat(l_3)\} \ .$$

The effects associated to the actions $pick(l)$ and $drop(l)$ in $O'$ are the support rules

$$pick(l) : K\neg hold, \ Kat(l) \ \rightarrow \ Khold \wedge K\neg at(l)$$
$$pick(l) : Khold \ \rightarrow \ K\neg hold \wedge Kat(l)$$
$$drop(l) : Khold \ \rightarrow \ K\neg hold \wedge Kat(l)$$

for each one of the three locations $l = l_i$, that condition each rule in $O$ with the empty tag, along with the support rules:

$$pick(l) : K\neg hold/at(l'), \ Kat(l)/at(l') \ \rightarrow \ Khold/at(l') \wedge K\neg at(l)/at(l')$$
$$pick(l) : Khold/at(l') \ \rightarrow \ K\neg hold/at(l') \wedge Kat(l)/at(l')$$
$$drop(l) : Khold/at(l') \ \rightarrow \ K\neg hold/at(l') \wedge Kat(l)/at(l')$$

that condition each rule in $O$ with the tags $at(l') \in T$, for $l' \in \{l_1, l_2\}$. The corresponding cancellation rules are:

$$pick(l) : \neg Khold, \ \neg K\neg at(l) \ \rightarrow \ \neg K\neg hold \wedge \neg Kat(l)$$
$$pick(l) : \neg K\neg hold \ \rightarrow \ \neg Khold \wedge \neg K\neg at(l)$$
$$drop(l) : \neg K\neg hold \ \rightarrow \ \neg Khold \wedge \neg K\neg at(l)$$

and

$$pick(l) : \neg Khold/at(l'), \ \neg K\neg at(l)/at(l') \ \rightarrow \ \neg K\neg hold/at(l') \wedge \neg Kat(l)/at(l')$$
$$pick(l) : \neg K\neg hold/at(l') \ \rightarrow \ \neg Khold/at(l') \wedge \neg K\neg at(l)/at(l')$$
$$drop(l) : \neg K\neg hold/at(l') \ \rightarrow \ \neg Khold/at(l') \wedge \neg K\neg at(l)/at(l') \ .$$

In addition, the actions in $O'$ include the merge actions $a_{m,hold}$ and $a_{m,at(l_3)}$ that follow from the merge $m = \{at(l_1), at(l_2)\}$ in $M$ for the literals *hold* and $at(l_3)$:

$$a_{m,hold} : Khold/at(l_1), Khold/at(l_2) \ \rightarrow \ Khold$$
$$a_{m,at(l_3)} : Kat(l_3)/at(l_1), Kat(l_3)/at(l_2) \ \rightarrow \ Kat(l_3) \ .$$

---

2. The empty tag is assumed in every $T$ and thus it is not mentioned explicitly.

It can be shown then that the plan

$$\pi_1' = \{pick(l_1), drop(l_3), pick(l_2), drop(l_3), a_{m,at(l_3)}\}$$

solves the classical problem $K_{T,M}(P)$ and hence, from Theorem 7, that the plan $\pi_1$ obtained from $\pi_1'$ by dropping the merge action, is a valid conformant plan for $P$ (shown above). We can see how some of the literals in $K_{T,M}(P)$ evolve as the actions in $\pi_1'$ are executed:

| | | |
|---|---|---|
| 0: | $Kat(l_1)/at(l_1), Kat(l_2)/at(l_2)$ | true in $I'$ |
| 1: | $Khold/at(l_1), Kat(l_2)/at(l_2)$ | true after $pick(l_1)$ |
| 2: | $Kat(l_3)/at(l_1), Kat(l_2)/at(l_2)$ | true after $drop(l_3)$ |
| 3: | $Kat(l_3)/at(l_1), Khold/at(l_2)$ | true after $pick(l_2)$ |
| 4: | $Kat(l_3)/at(l_1), Kat(l_3)/at(l_2)$ | true after $drop(l_3)$ |
| 5: | $Kat(l_3)$ | true after merge $a_{m,at(l_3)}$. |

We can also verify in the same manner that the action sequence $\pi_2'$

$$\pi_2' = \{pick(l_1), pick(l_2), a_{m,hold}, drop(l_3)\}$$

is not a classical plan for $K_{T,M}(P)$, the reason being that the atom $Khold/at(l_1)$ holds after the first pick up action but not after the second. This is due to the cancellation rule:

$$pick(l_2) : \neg K\neg hold/at(l_1) \rightarrow \neg Khold/at(l_1) \ \wedge \ \neg K\neg at(l_2)/at(l_1)$$

that expresses that under the assumption $at(l_1)$ in the initial situation, $hold$ and $\neg at(l_2)$ are not known to be true after the action $pick(l_2)$, if under the same assumption, $\neg hold$ was not known to be true before the action.

## 5. A Complete Translation: $K_{S0}$

A *complete* instance of the translation scheme $K_{T,M}$ can be obtained in a simple manner by setting the tags to the possible initial states of the problem $P$ and by having a merge for each precondition and goal literal $L$ that includes all these tags. We call the resulting 'exhaustive' translation $K_{S0}$:

**Definition 8** (Translation $K_{S0}$). For a conformant problem $P$, the translation $K_{S0}(P)$ is an instance of the translation $K_{T,M}(P)$ where

- $T$ is set to the union of the empty tag and the set $S_0$ of all possible initial states of $P$ (understood as the maximal sets of literals that are consistent with $I$), and

- $M$ is set to contain a single merge $m = S_0$ for each precondition and goal literal $L$ in $P$.

The translation $K_{S0}$ is valid and hence sound, and it is complete due the correspondence between tags and possible initial states:

**Theorem 9** (Completeness of $K_{S0}$). *If $\pi$ is a conformant plan for $P$, then there is a classical plan $\pi'$ for $K_{S0}(P)$ such that $\pi$ is the result of dropping the merge actions from $\pi'$.*

| Problem | $\#S_0$ | $K_{S0}$ | | POND | | CFF | |
|---|---|---|---|---|---|---|---|
| | | time | len | time | len | time | len |
| adder-01 | 18 | $> 2h$ | | 0,4 | 26 | $> 2h$ | |
| blocks-02 | 18 | 0,2 | 23 | 0,4 | 26 | $> 2h$ | |
| blocks-03 | 231 | 59,2 | 80 | 126,8 | 129 | $> 2h$ | |
| bomb-10-1 | 1k | 5,9 | 19 | 1 | 19 | 0 | 19 |
| bomb-10-5 | 1k | 11,3 | 15 | 3 | 15 | 0 | 15 |
| bomb-10-10 | 1k | 18,3 | 10 | 8 | 10 | 0 | 10 |
| bomb-20-1 | 1M | $> 2.1GB$ | | 4139 | 39 | 0 | 39 |
| coins-08 | 1k | 20,2 | 27 | 2 | 28 | 0 | 28 |
| coins-09 | 1k | 19,9 | 25 | 5 | 26 | 0 | 26 |
| coins-10 | 1k | 21,5 | 31 | 5 | 28 | 0,1 | 38 |
| coins-11 | 1M | $> 2.1GB$ | | $> 2h$ | | 1 | 78 |
| comm-08 | 512 | 18,3 | 61 | 1 | 53 | 0 | 53 |
| comm-09 | 1k | 77,7 | 68 | 1 | 59 | 0 | 59 |
| comm-10 | 2k | $> 2.1GB$ | | 1 | 65 | 0 | 65 |
| corners-square-16 | 4 | 0,2 | 102 | 1131 | 67 | 13,1 | 140 |
| corners-square-24 | 4 | 0,7 | 202 | $> 2h$ | | 321 | 304 |
| corners-square-28 | 4 | 1,2 | 264 | $> 2h$ | | $> 2h$ | |
| corners-square-116 | 4 | 581,4 | 3652 | $> 2h$ | | $> 2h$ | |
| corners-square-120 | 4 | $> 2.1GB$ | | $> 2h$ | | $> 2h$ | |
| square-center-16 | 256 | 13,1 | 102 | 1322 | 61 | $> 2h$ | |
| square-center-24 | 576 | $> 2.1GB$ | | $> 2h$ | | $> 2h$ | |
| log-2-10-10 | 1k | 183,5 | 85 | $> 2h$ | | 1,6 | 83 |
| log-3-10-10 | 59k | $> 2h$ | | $> 2h$ | | 4,7 | 108 |
| ring-5 | 1,2k | 12,6 | 17 | 6 | 20 | 4,3 | 31 |
| ring-6 | 4,3k | $> 2.1GB$ | | 33 | 27 | 93,6 | 48 |
| safe-50 | 50 | 0,5 | 50 | 9 | 50 | 29,4 | 50 |
| safe-70 | 70 | 1,4 | 70 | 41 | 70 | 109,9 | 70 |
| safe-100 | 100 | 6 | 100 | $> 2.1GB$ | | 1252,4 | 100 |
| sortnet-07 | 256 | 2,9 | 28 | 480 | 25 | SNH | |
| sortnet-08 | 512 | 9,8 | 36 | $> 2h$ | | SNH | |
| sortnet-09 | 1k | 77,7 | 45 | $> 2h$ | | SNH | |
| sortnet-10 | 2k | $> 2.1GB$ | | $> 2h$ | | SNH | |
| uts-k-08 | 16 | 0,6 | 46 | 24 | 47 | 4,4 | 46 |
| uts-k-10 | 20 | 1,2 | 58 | 2219 | 67 | 16,5 | 58 |

Table 1: $K_{S0}$ translation fed into FF planner compared with POND and Conformant FF (CFF) along both times and reported plan lengths. $\#S_0$ stands for number of initial states, 'SNH' means goal syntax not handled (by CFF). Times reported in seconds and rounded to the closest decimal.

For problems $P$ whose actions have no preconditions, the argument is simple: if $\pi$ is a conformant plan for $P$ then $\pi$ must be a classical plan for $P/s$ for each possible initial state $s$, but then if $\pi$ achieves the (goal) literal $G_i$ in $P/s$ for each $s$, $\pi$ must achieve the literal $KG_i/s$ in $K_{S0}(P)$ for each $s$ as well, so that $\pi$ followed by the merge action for $G_i$, must achieve the literal $KG_i$. In the presence of action preconditions, this argument must be applied inductively on the plan length, but the idea remains the same (see the proof in the appendix for details): a correspondence can be established between the evolution of the fluents $L$ in each problem $P/s$ and the evolution of the fluents $KL/s$ in the problem $K_{S0}(P)$.

The significance of the exhaustive $K_{S0}$ translation is not only theoretical. There are plenty of conformant problems that are quite hard for current planners even if they involve a handful of possible initial states. An example of this is the Square-Center-$n$ task (Cimatti et al., 2004), where an agent has to reach the center of an empty square grid with certainty, not knowing its initial location. There are four actions that move the agent one unit in each direction, except when in the border of the grid, where they have no effects. In the standard version of the problem, the initial position is fully unknown resulting in $n^2$ possible initial states, yet the problem remains difficult, and actually beyond the reach of most planners, for small values of $n$, even when the uncertainty is reduced to *a pair of possible initial states*. The reason is that the agent must locate itself before heading for the goal. The domain Corners-Square-$n$ in Table 1 is a variation of Square-Center-$n$ where the possible initial states are the four corners of the grid.

Table 1 shows results for a conformant planner based on the $K_{S0}(P)$ translation that uses FF (Hoffmann & Nebel, 2001) for solving the resulting classical problem, and compares it with two of the planners that entered the Conformant track of the 2006 Int. Planning Competition (Bonet & Givan, 2006): POND (Bryce et al., 2006) and Conformant FF (Hoffmann & Brafman, 2006) (the other two planners in the competition were translation-based: $T_0$, based on the formulation developed in this paper, and $K(P)$, based on an earlier and more restricted formulation due to Palacios & Geffner, 2006). Clearly, the approach based on the $K_{S0}(P)$ translation does not scale up to problems with many possible initial states, yet when the number of such states is small, it does quite well.

## 6. Complete Translations that May be Compact Too

In order to have complete translations that are polynomial, certain assumptions about the formulas in the initial situation $I$ need to be made. Otherwise, just checking whether a goal is true in $I$ is intractable by itself, and therefore a polynomial but complete translation would be impossible (unless P = NP). We will thus assume that $I$ is in *prime implicate (PI) form* (Marquis, 2000), meaning that $I$ includes only the inclusion-minimal clauses that it entails but no tautologies. It is known that checking whether a clause follows logically from a formula $I$ in PI form reduces to checking whether the clause is subsumed by a clause in $I$ or is a tautology, and hence is a polynomial operation. The initial situations $I$ in most benchmarks is in $PI$ form or can easily be cast into PI form as they are normally specified by means of a set of non-overlapping $oneof(X_1, \ldots, X_n)$ expressions that translate into clauses $X_1 \vee \cdots \vee X_n$ and binary clauses $\neg X_i \vee \neg X_j$ for $i \neq j$ where any resolvent is a tautology.

## 6.1 Conformant Relevance

The translation $K_{S0}(P)$ is complete but introduces a number of literals $KL/t$ that is exponential in the worst case: one for each possible initial state $s_0$. This raises the question: is it possible to have complete translations that are not exhaustive in this sense? The answer is yes and in this section we provide a simple condition that ensures that a translation $K_{T,M}(P)$ is complete. It makes use of the notion of relevance:[3]

**Definition 10** (Relevance). The conformant relevance relation $L \longrightarrow L'$ in $P$, read $L$ is relevant to $L'$, is defined inductively as

1. $L \longrightarrow L$
2. $L \longrightarrow L'$ if $a : C \rightarrow L'$ is in $P$ with $L \in C$ for some action $a$ in $P$
3. $L \longrightarrow L'$ if $L \longrightarrow L''$ and $L'' \longrightarrow L'$
4. $L \longrightarrow L'$ if $L \longrightarrow \neg L''$ and $L'' \longrightarrow \neg L'$.

The first clause stands for reflexivity, the third for transitivity, the second captures conditions that are relevant to the effect, and the fourth, the conditions under which $L$ preempts conditional effects that may delete $L'$. If we replace 4 by

4' $L \longrightarrow L'$ if $\neg L \rightarrow \neg L'$

which is equivalent to 4 in the context of 1–3, the resulting definition is the one by Son and Tu (2006), where the notion of relevance is used to generate a limited set of possible 'partial' initial states over which the 0-approximation is complete (see Section 11 for a discussion on the relation between tags and partial initial states).

Notice that according to the definition, a precondition $p$ of an action $a$ is not taken to be 'relevant' to an effect $q$. The reason is that we want the relation $L \longrightarrow L'$ to capture the conditions under which *uncertainty about $L$* is relevant *to the uncertainty about $L'$*. This is why we say this is a relation of *conformant relevance*. Preconditions must be known to be true in order for an action to be applied, so they do not introduce nor propagate uncertainty into the effects of an action.

If we let $C_I$ stand for the set of clauses representing uncertainty about the initial situation, namely, the non-unit clauses in $I$ along with the tautologies $L \vee \neg L$ for complementary literals $L$ and $\neg L$ not appearing as unit clauses in $I$, the notion of (conformant) relevance can be extended to clauses as follows:

**Definition 11** (Relevant Clauses). A clause $c \in C_I$ is relevant to a literal $L$ in $P$ if all literals $L' \in c$ are relevant to $L$. The set of clauses in $C_I$ relevant to $L$ is denoted as $C_I(L)$.

Having a representation of the uncertainty in the initial situation that is relevant to a literal $L$, it is possible to analyze the completeness of a translation $K_{T,M}$ in terms of the relation between the merges $m$ for the literals $L$, on one hand, and the sets of clauses $C_I(L)$ that are relevant to $L$ on the other.

---

3. While we follow an earlier account (Palacios & Geffner, 2007), many of the definitions and theorems differ in a number of details (for example, the notion of relevance depends on the rules in $P$ but not on the clauses in the initial situation). The changes are aimed at making the resulting formulation simpler and cleaner.

## 6.2 Covering Translations

It may appear that a translation $K_{T,M}$ would be complete when the merges $m$ for precondition and goal literals $L$, understood as the DNF formulas $\bigvee_{t \in m} t$, contain as much information, and thus are equivalent to the CNF formula $C_I(L)$ that captures the fragment of the initial situation $I$ that is relevant to $L$. This intuition is partially correct, but misses one important point; namely that not every DNF formula equivalent to $C_I(L)$ will do: the DNF representation captured by the merges must be 'vivid' enough. For example, if $C_I(L)$ is the single clause $x \vee \neg x$, completeness requires a tag for $x$, a tag for $\neg x$, and a merge $m = \{x, \neg x\}$ for $L$ containing the two tags, even if the clause $x \vee \neg x$ is a tautology and is thus equivalent to the DNF formula *true*.

For defining the types of tags and merges that are required for completeness then, let us first define the *closure* $S^*$ of a set of literals $S$, relative to a conformant problem $P = \langle F, I, O, G \rangle$, as the set of literals that follow from $S$ and $I$:

$$S^* = \{ L \mid I, S \models L \} \ .$$

Let us also say that $S$ is *consistent* if $S^*$ does not contain a pair of complementary literals.

The type of merges $m$ required for precondition and goal literals $L$ are then those that do not only imply $C_I(L)$ but that *satisfy* it as well. The notion of satisfaction associates a consistent set of literals $S$ with the *partial truth assignment* that is implicit in the closure $S^*$ of $S$, and is extended to account for the conditions under which a DNF formula (e.g., a merge for $L$) satisfies a CNF formula (e.g., $C_I(L)$).

**Definition 12** (Satisfaction).     1. A consistent set of literals $S$ *satisfies a clause* $L_1 \vee L_2 \vee \cdots \vee L_m$ if $S^*$ contains one of the literals $L_i$, $i = 1, \ldots, m$.

    2. A consistent set of literals $S$ *satisfies a collection of clauses* $\mathcal{C}$ if $S$ satisfies each clause in $\mathcal{C}$.

    3. A collection $\mathcal{S}$ of consistent sets of literals *satisfies* a collection of clauses $\mathcal{C}$ if each set $S$ in $\mathcal{S}$ satisfies $\mathcal{C}$.

The type of merges required for completeness are then simply the valid merges $m$ that satisfy the set of clauses $C_I(L)$. We call them *covering merges*:

**Definition 13** (Covering Merges). A valid merge $m$ in a translation $K_{T,M}(P)$ *covers* a literal $L$ if $m$ satisfies $C_I(L)$.

For example, if $C_I(L)$ is given by the clauses that result from a *oneof*$(x_1, \ldots, x_n)$ expression, i.e. $x_1 \vee x_2 \vee \cdots \vee x_n$ and $\neg x_i \vee \neg x_j$ for all $i$ and $j$, $1 \leq i, j \leq n$, $i \neq j$, then the merge $m = \{x_1, \ldots, x_n\}$ covers the literal $L$, as each $x_i^*$ not only includes $x_i$ but also $\neg x_j$ for all $j \neq i$, and thus $x_i^*$ satisfies $C_I(L)$.

If for a merge $m = \{t_1, \ldots, t_n\}$, we denote by $m^*$ the DNF formula $\bigvee_{t_i \in m} t_i^*$, where each tag $t_i$ is replaced by its closure $t_i^*$, then it is simple to prove that if $m$ covers the literal $L$, $m^*$ entails $C_I(L)$. A merge $m$ that covers $L$ is thus a DNF formula that is strong enough to imply the CNF formula $C_I(L)$ (through the closure), weak enough to be entailed by $I$, and vivid enough to satisfy $C_I(L)$.

As a further illustration, if $C_I(L)$ is given by the tautologies $p \vee \neg p$ and $q \vee \neg q$, and $I = C_I(L)$, the merge $m_1 = \{p, \neg p\}$ implies $C_I(L)$ but does not satisfy $C_I(L)$. Likewise, the merge $m_2 = \{\{p, q\}, \{\neg p, \neg q\}\}$ satisfies $C_I(L)$ but is not entailed by $I$. Finally, the merge $m_3 = \{\{p, q\}, \{p, \neg q\}, \{\neg p, q\}, \{\neg p, \neg q\}\}$ satisfies $C_I(L)$ and is entailed by $I$, and thus is a valid merge that covers $L$.

If a valid translation $K_{T,M}(P)$ contains a merge $m$ that covers $L$ for each precondition and goal literal $L$ in $P$, we say that the translation *covers* $P$ or just that it is a *covering translation*:

**Definition 14** (Covering Translation). A covering translation is a valid translation $K_{T,M}(P)$ that includes one merge that covers $L$, for each precondition and goal literal $L$ in $P$.

A central result of the paper is that covering translations are complete:

**Theorem 15** (Completeness). *Covering translations $K_{T,M}(P)$ are complete; i.e., if $\pi$ is a conformant plan for $P$, then there is a classical plan $\pi'$ for $K_{T,M}(P)$ such that $\pi$ is $\pi'$ with the merge actions removed.*

In other words, complete translations $K_{T,M}(P)$ result when the tags and merges in $T$ and $M$ capture the information in the initial situation that is relevant to each precondition and goal literal in a suitable manner.

Theorem 15 can be used in two ways: for proving the completeness of a translation, by checking that the covering condition holds, and for constructing complete translations, by enforcing the covering condition. In addition, while our interest in this paper is on conformant planning with no optimality guarantees, the theorem is useful for *optimal conformant planning* as well, whether the cost of plans is defined as their length (action costs equal to 1) or as the sum of non-uniform action costs. In both cases, the theorem ensures that the problem of optimal conformant planning gets mapped into a problem of optimal classical planning provided that the cost of the merge actions in $K_{T,M}(P)$ is made sufficiently small.

As an illustration of Theorem 15, consider the conformant problem $P$ with initial situation $I = \{x_1 \vee \cdots \vee x_m\}$, goal $G = L$, and actions $a_i, i = 1, \ldots, m$, each with effect $x_i \rightarrow L$. The number of possible initial states for this problem is exponential in $m$, as the disjunction among the $x_i$'s is not exclusive. So, the translation $K_{S0}(P)$ is complete but exponential in size. On the other hand, consider the translation $K_{T,M}(P)$ where $T = \{x_1, \ldots, x_m\}$ and $M$ contains the single valid merge $m = \{x_1, \ldots, x_m\}$ for $L$. It is simple to verify that this merge covers the goal $L$ (satisfies $C_I(L) = I$), and hence that the translation $K_{T,M}(P)$ is covering, and by Theorem 15, complete, while being polynomial in $m$.

Notice that testing whether a valid translation $K_{T,M}(P)$ is a covering translation can be done in polynomial time, as in particular, computing the set of literals $t^*$ from every tag $t$ in $T$ is a tractable operation provided that $I$ is in PI form; indeed, $I, t \models L'$ iff $I \models t \supset L'$ iff $\neg t \vee L'$ is a tautology or is subsumed by a clause in $I$.

## 6.3 Translation *Kmodels*

It is straightforward to show that the exponential translation $K_{S0}$ considered in Section 3, where (non-empty) tags stand for the possible initial states, is covering and hence complete according to Theorem 15. It is possible, however, to take further advantage of Theorem 15 for devising a complete translation that is usually more compact. We call it *Kmodels*.

**Definition 16.** The translation $Kmodels(P)$ is obtained from the general scheme $K_{T,M}(P)$ by defining

- $M$ to contain one merge $m$ for each precondition and goal literal $L$ given by the models of $C_I(L)$ that are consistent with $I$,[4] and

- $T$ to contain the tags in all such merges along with the empty tag.

The translation *Kmodels* is equivalent to $K_{S0}$ when for all the precondition and goal literals $L$, $C_I(L) = I$; i.e., when all the clauses in $I$ are relevant to $L$. Yet, in other cases, the first translation is exponential in the number of variables appearing in one such $C_I(L)$ set (the one with the largest number of such variables), while the second is exponential in the number of unknown variables in $I$. For example, if there are $n$ precondition and goal literals $L_i$, $i = 1, \ldots, n$ in $P$ such that for each one, $C_I(L_i)$ is a unique $oneof(x^i_1, \ldots, x^i_m)$ expression, the merge for the literal $L_i$ in $K_{S0}(P)$ will contain the $m^n$ models of the $n$ one-of expressions in $I$, while the merge for $L_i$ in $Kmodels(P)$ will just contain the $m$ models of the single $oneof(x^i_1, \ldots, x^i_m)$ expression in $C_I(L_i)$. The translation *Kmodels* can thus be exponentially more compact than the exhaustive $K_{S0}$ translation while remaining sound and complete:

**Theorem 17.** *The translation Kmodels(P) is sound and complete.*

In the worst case, however, *Kmodels* is also an exponential translation. We thus consider next *polynomial* translations and the conditions under which they are complete.

## 6.4 Conformant Width

We address now the conditions under which a compact, covering translation can be constructed in polynomial time. For this, we define a structural parameter that we call the *conformant width* of a problem $P$, that in analogy to the notion of width used in graphical models (Dechter, 2003), will provide an upper bound on the time and space complexity required for generating a covering translation. More precisely, the complexity of this construction will be exponential in the conformant width of the problem $P$ that cannot exceed the number of fluents in $P$ but can be much lower.

In principle, we would like to define the width $w(P)$ as the maximum tag size required in a translation $K_{T,M}(P)$ to be a covering translation. Such a definition, however, would not give us the complexity bounds that we want, as just checking the validity of a merge with tags of bounded size is an intractable operation, whether the initial situation $I$ is in prime implicate form or not.[5] So we need to define width in a different way. First, let the *cover* of a set of clauses be defined as follows:

---

4. The models of $C_I(L)$ are to be understood as conjuntions of literals.

5. The problem of checking whether $I$ entails a DNF formula whose terms may have more than 2 literals is coNP-hard even if $I$ is equivalent to true. Indeed, if $\Phi$ is a 3-CNF formula; $\Phi$ is contradictory iff its negation $\neg\Phi$ (which is in 3-DNF) is valid, which in turn is true iff $\neg\Phi$ is implied by $I$. Actually, for a general $I$ in prime implicate form, the problem remains coNP-hard even if the terms of the DNF formula contain at most 2 literals. We thank Pierre Marquis for pointing these results to us.

**Definition 18** (Cover). The cover $c(\mathcal{C})$ of a set of clauses $\mathcal{C}$, relative to a conformant problem $P$ with initial situation $I$, is the collection of all minimal sets of literals $S$ consistent with $I$ such that $S$ contains a literal of each clause in $\mathcal{C}$.

Two important properties of the cover $c(\mathcal{C})$ of a set of clauses $\mathcal{C}$ are that $c(\mathcal{C})$ stands for a DNF formula that is logically equivalent to the CNF formula $\mathcal{C}$ given $I$, and that $c(\mathcal{C})$ can be computed in polynomial time if the size of $\mathcal{C}$ is bounded by a constant. Moreover, $c(\mathcal{C})$ not only implies $\mathcal{C}$ but *satisfies* $\mathcal{C}$ as well. Thus in particular, if $\mathcal{C}$ is the collection of clauses $C_I(L)$ that are relevant to the literal $L$, the cover $c(C_I(L))$ of $C_I(L)$ is a valid merge that covers $L$. From this and the completeness of covering translations, it follows that a complete translation $K_{T,M}(P)$ can be constructed in polynomial time if the size $|C_I(L)|$ of the sets of clauses $C_I(L)$ for all precondition and goal literals $L$ in $P$ is bounded. Unfortunately, this condition rarely seems to hold, yet there is a weaker sufficient condition that does: namely, it is often possible to find a subset $\mathcal{C}$ of clauses that are either in $C_I(L)$ or are tautologies such that $c(\mathcal{C})$ satisfies $C_I(L)$ and thus covers the literal $L$. We thus define the *width of the literal $L$* as the size of the smallest such set (cardinality-wise). For this, we denote by $C_I^*(L)$ the set of clauses $C_I(L)$ extended with tautologies of the form $p \vee \neg p$ for fluents $p$ such that either $p$ or $\neg p$ appears in $C_I(L)$ (if both appear in $C_I(L)$ then $p \vee \neg p$ is in $C_I(L)$ from its definition).

**Definition 19** (Width of Literal). The conformant width of a literal $L$ in $P$, written $w(L)$, is the size of the smallest (cardinality-wise) set of clauses $\mathcal{C}$ in $C_I^*(L)$ such that $c(\mathcal{C})$ satisfies $C_I(L)$.

A consequence of this definition is that the width of a literal must lie in the interval $0 \leq w(L) \leq n$, where $n$ is the number of fluents in $P$ whose status in the initial situation is not known. Indeed, if $C_I(L)$ is empty, $w(L) = 0$, while for any set of clauses $C_I(L)$, the cover $c(\mathcal{C})$ of the set $\mathcal{C}$ of tautologies in $C_I^*(L)$ must satisfy $C_I(L)$, and thus $w(L) \leq |\mathcal{C}| \leq n$. Similarly, if $C_I(L)$ contains a single clause $x_1 \vee \cdots \vee x_m$ or the clauses $x_1 \vee \cdots \vee x_m$ and $\neg x_i \vee \neg x_j$ that correspond to the *oneof*$(x_1, \ldots, x_m)$ expression, it is simple to prove that $w(L) = 1$ with the singleton $\mathcal{C} = \{x_1 \vee \cdots \vee x_m\}$ generating the cover $c(\mathcal{C}) = \{\{x_1\}, \ldots, \{x_n\}\}$ that satisfies $C_I(L)$. Finally, if $C_I(L)$ contains the two tautologies $p \vee \neg p$ and $q \vee \neg q$, $w(L) = 2$ as the smallest $\mathcal{C}$ in $C_I^*(L)$ whose cover satisfies $C_I(L)$ is $C_I(L)$ itself.

The width of a problem is the width of the precondition or goal literal with maximum width:

**Definition 20** (Width of Problem). The conformant width of a problem $P$, written as $w(P)$, is $w(P) = \max_L w(L)$, where $L$ ranges over the precondition and goal literals in $P$.

We show below that for problems with bounded width, complete translations can be constructed in polynomial time, and moreover, that almost all existing conformant benchmarks have bounded width, and more precisely, width equal to 1. In such a case, the resulting translations will use tags that are never greater in size than $w(P)$, so that for problems with width 1, tags will be single literals.

Like for the (tree)width of graphical models, computing the width of a problem $P$ is exponential in $w(P)$, so the recognition of problems with small width can be carried out quite efficiently:

**Proposition 21** (Determining Width). *The width $w(P)$ of $P$ can be determined in time that is exponential in $w(P)$.*

In particular, we can test if $w(P) = 1$ by considering one by one each of the sets $\mathcal{C}$ that includes a single clause from $C_I^*(L)$, verifying whether $c(\mathcal{C})$ satisfies $C_I(L)$ or not. If $w(P) \nleq 1$, then the same verification must be carried out by setting $\mathcal{C}$ to each set of $i$ clauses in $C_I^*(L)$ for increasing values of $i$. For a fixed value of $i$, there is a polynomial number of such clause sets $\mathcal{C}$ and the verification of each one can be done in polynomial time. Moreover, from the arguments above regarding $w(L)$, $w(P)$ can never exceed the number of unknown fluents in the problem:

**Proposition 22** (Bounds on Width). *The width of $P$ is such that $0 \leq w(P) \leq n$, where $n$ is the number of fluents whose value in the initial situation is not known.*

### 6.5 Polynomial Translation $K_i$

The translation $K_i$, where the parameter $i$ is a non-negative integer, is an instance of the general $K_{T,M}$ scheme designed to be sound, polynomial for a fixed $i$, and complete for problems with width $w(P) \leq i$. Thus, for example, the translation $K_1$ is sound, polynomial, and complete for problems with width 1.

**Definition 23** (Translation $K_i$). The translation $K_i(P)$ is obtained from the general scheme $K_{T,M}(P)$ where

- $M$ is set to contain one merge $m = c(\mathcal{C})$ for each precondition and goal literal $L$ in $P$ if there is a set $\mathcal{C}$ of at most $i$ clauses in $C_I^*(L)$ such that $m$ covers $L$. If no such set exists, one merge $m = c(\mathcal{C})$ for $L$ is created for each set $\mathcal{C}$ of $i$ clauses in $C_I^*(L)$, and no merges are created for $L$ if $C_I^*(L)$ is empty;

- $T$ is the collection of tags appearing in those merges and the empty tag.

The translation $K_i(P)$ applies to problems $P$ of any width, remaining in all cases exponential in $i$ but polynomial in the number of fluents, actions, and clauses in $P$. In addition, the translation $K_i(P)$ is sound, and for problems with width bounded by $i$, complete.

**Theorem 24** (Properties $K_i$). *For a fixed $i$, the translation $K_i(P)$ is sound, polynomial, and if $w(P) \leq i$, covering and complete.*

Soundness is the result of the merges being all valid by construction, as the covers $c(\mathcal{C})$ for any $\mathcal{C}$ in $C_I^*(L)$ are entailed by $\mathcal{C}$ and hence by $I$. The complexity is polynomial for a fixed $i$, because there is a polynomial number of clause sets $\mathcal{C}$ of size $i$ in $C_I^*(L)$, and constructing the cover $c(\mathcal{C})$ for each one of them, is a polynomial operation. Finally, completeness follows from the definition of width: if $w(P) \leq i$, then there is a set of clauses $\mathcal{C}$ in $C_I^*(L)$ with size $|\mathcal{C}|$ no greater than $i$ whose cover satisfies $C_I(L)$, and thus $M$ in $K_i(P)$ must contain a merge $m = c(\mathcal{C})$ for $L$ that covers $L$.

Notice that for $i = 0$, the translation $K_i(P)$ reduces to the basic $K_0(P)$ translation introduced in Section 3 that has no tags (other than the empty tag) and no merges. Before, we assessed the completeness of this translation in terms of the 0-approximation semantics. Theorem 24 provides an alternative interpretation: the translation $K_0(P)$ is complete for problems $P$ with zero width. These are the problems for which the set of clauses $C_I(L)$ relevant to a precondition or goal literal $L$ is empty. This makes precise the intuition mentioned above that the $K_0(P)$ translation is complete for problems where the uncertain information in $I$ is not relevant. In such cases, none of the clauses in the initial situation $I$ make it into the sets of relevant clauses $C_I(L)$ for preconditions and goal literals $L$.

| | Domain-Parameter | # Unknown Fluents | Width |
|---|---|---|---|
| 1 | Safe-$n$ combinations | $n$ | 1 |
| 2 | UTS-$n$ locs | $n$ | 1 |
| 3 | Ring-$n$ rooms | $4n$ | 1 |
| 4 | Bomb-in-the-toilet-$n$ bombs | $n$ | 1 |
| 5 | Comm-$n$ signals | $n$ | 1 |
| 6 | Square-Center-$n \times n$ grid | $2n$ | 1 |
| 7 | Cube-Center-$n \times n \times n$ cube | $3n$ | 1 |
| 8 | Grid-$n$ shapes of $n$ keys | $n \times m$ | 1 |
| 9 | Logistics $n$ pack $m$ locs | $n \times m$ | 1 |
| 10 | Coins-$n$ coins $m$ locs | $n \times m$ | 1 |
| 11 | Block-Tower-$n$ Blocks | $n \times (n-1) + 3n + 1$ | $n \times (n-1) + 3n + 1$ |
| 12 | Sortnet-$n$ bits | $n$ | $n$ |
| 13 | Adder $n$ pairs of bits | $2n$ | $2n$ |
| 14 | Look-and-Grab $m$ objs from $n \times n$ locs | $n \times n \times m$ | $m$ |
| 15 | 1-dispose $m$ objs from $n \times n$ locs | $n \times n \times m$ | $m$ |

Table 2: Width of parameterized domains

As an illustration of Theorem 24, consider again the conformant problem $P$ with initial situation $I = \{x_1 \vee \cdots \vee x_m\}$, goal $G = \{L\}$, and actions $a_i, i = 1, \ldots, m$, each with effect $x_i \to L$. For this problem, the singleton set of clauses $\mathcal{C} = C_I(L) = I$ is such that $c(\mathcal{C}) = \{\{x_1\}, \ldots, \{x_m\}\}$ covers $C_I(L)$. Then, since there is no other precondition or goal literal, $K_1(P)$ includes the single merge $m = c(\mathcal{C})$ for $L$ with the singleton tags $t_i = \{x_i\}$, that we write simply as $m = \{x_1, \ldots, x_m\}$. The translation $K_1(P)$ is polynomial in $m$, and since $w(P) = 1$, by Theorem 24 it is complete. Notice that for this same example, the translations $K_{S0}(P)$ and $Kmodels(P)$ are identical and exponential in $m$ (the number of models of $I$ and $C_I(L)$).

## 6.6 Width of Conformant Benchmarks

The practical value of the notion of width becomes apparent when the width of existing benchmarks is considered. Table 2 summarizes the width of many of the existing benchmark domains for conformant planning. The domains all depend on certain parameters $n$ or $m$ that capture the size of the instances (e.g., size of a grid, number of objects, etc.).[6] A *domain* has a bounded width when its width does not grow with the size of its instances, and has width equal to $i$ when all of its instances have width $i$ regardless of the parameter values.

As it can be seen from the table, the width of most existing benchmarks is 1. In all these cases, this means that the sets $C_I(L)$ of clauses that are relevant to a precondition or goal literal $L$ contain a single clause (often a tautology $p \vee \neg p$ or a disjunction $x_1 \vee \ldots \vee x_m$) or a single $oneof(x_1, \ldots, x_m)$ expression (that translates into the disjunction $x_1 \vee \cdots \vee x_m$ and clauses $\neg x_i \vee \neg x_k$). As shown above, $w(L)$, and therefore, $w(P)$, is equal to 1 in theses cases.

---

6. The names of the parameterized domains in the table do not coincide with the names of the instances as currently used. E.g. Comm-$n$ in IPC5 refers to a Communication instance but not necessarily to an instance with $n$ signals.

On the other extreme are domains such as Blocks, Sortnet, and Adder, all of which have maximal widths; i.e., widths that are equivalent to the number of fluents whose status in the initial situation is not known. This is because all fluents interact through the action conditions (not the preconditions). The numbers for Blocks in Table 2, thus follow from the number of fluents involved; namely, the fluents $on(x,y)$, $clear(x)$, $ontable(x)$, and $holding(x)$.

Finally, the domains 1-dispose and Look-and-Grab (Palacios & Geffner, 2006, 2007) where $m$ objects with unknown locations in a grid of $n$ by $n$ must be collected by a robot whose gripper can hold one object at a time, have width equal to $m$, meaning that the width of these domains grows with the number of objects but not with the size of the grid. This is because in this case, the clauses about the possible locations of the $m$ objects are all relevant to the condition 'hand empty' of the pick up actions.

Let us point out that the completeness of the translation $K_i(P)$ for problems $P$ with width $w(P)$ bounded by $i$, establishes a correspondence between the conformant plans for $P$ and the classical plans for $K_{T,M}(P)$. For solving $P$, however, this correspondence is not needed; it suffices for $K_i(P)$ to be *solvable*; a plan for $K_i(P)$ will then encode a conformant plan for $P$, even if $K_i(P)$ does not capture *all* conformant plans for $P$. From this perspective, it makes sense to refer to the smallest value of the $i$ parameter for which the classical problem $K_i(P)$ is solvable, as the *effective width* of $P$, denoted $w_e(P)$. It turns out that while $w_e(P)$ cannot be larger than $w(P)$, it may be much smaller. An interesting example of this comes from the Sortnet-$n$ domain (Bonet & Geffner, 2000). Sortnet-$n$ is considered a challenging domain in conformant planning with very few planners able to scale up to even small values of $n$ (the number of entries to be sorted in a sorting network). The domain has width $n$, and in the compact encoding used in IPC5, the input vector is represented by a set of bits, exploiting the fact that sorting vectors of numbers reduces to sorting vector of bits (0's and 1's). The domain cannot be solved by the $K_1$ translation that FF reports correctly as unsolvable after a brief unsuccessful search. On the other hand, it is possible to reformulate the domain, replacing the unary $high(i)$ and $low(i)$ predicates by binary predicates $less(i,j)$ that compare two vector entries. We call this reformulation Sort-2-$n$. While the encoding Sort-$n$ is linear in $n$, the encoding Sort-2-$n$ is quadratic in $n$, and in both cases, the problem width is maximum, given by the number of fluents whose status in the initial situation is unknown. Yet, while the more compact Sort-$n$ encoding is not solvable by the $K_1$ translation, $K_1$ suffices to solve the problem over the expanded Sort-2-$n$ encoding that actually can also be solved by $K_0$. Thus the effective width of Sort-2-$n$ is 0. Interestingly, provided the $K_0$ translation of Sort-2-$n$, instances can be solved with up to 20 entries. On the other hand, conformant planners such as Conformant-FF and POND can solve Sort-2-$n$ instances for $n$ no greater than 3.

## 7. Tags and Initial States

A deeper understanding of the results above can be obtained by relating tags with possible initial states. By looking more closely at this relation in the context of covering translations, we will be able to answer the question of how a polynomial number of contexts (tags) can play the role of an exponential number of possible initial states in problems with bounded width.

For this, let us first recall a notation introduced in Section 2.2, where for a state $s$, we wrote $I(s)$ to refer to the set of atoms encoding $s$ (i.e, $p \in I(s)$ iff $p$ is true in $s$) and $P/s$ to refer to the *classical* planning problem $P/s = \langle F, I(s), O, G \rangle$ that is like the conformant problem $P = \langle F, I, O, G \rangle$ but with the initial state fixed to $s$.

Let us now extend this notation and say that an action sequence $\pi$ *conforms* with a set of states $S$ given the conformant problem $P$ iff $\pi$ is a plan for the classical problem $P/s$ for each $s \in S$. Clearly, a conformant plan for $P$ is nothing else but an action sequence that conforms with the set $S_0$ of possible initial states of $P$, yet the notion of 'conforms' allows us to abstract away the initial situation $I$ and make precise the notion of a *basis*:

**Definition 25** (Basis for $P$). A set of states $S'$ is a *basis* for a conformant problem $P = \langle F, I, O, G \rangle$ if $S'$ is a subset of the set $S_0$ of possible initial states of $P$ and every plan that conforms with $S'$ conforms with the set $S_0$ of possible initial states $S_0$.

In words, if $S'$ is a basis for $P$, it is not necessary to consider all the states in $S_0$ for computing the conformant plans for $P$; it suffices to consider just the states in $S'$. We aim to show that if the width of $P$ is bounded, then $P$ has a polynomial basis $S'$ even if $S_0$ has exponential size. Moreover, the states $s$ in such a basis are in close correspondence with the tags appearing in a covering translation.

As an illustration, consider a problem $P$ with actions $a_i$, $i = 1, \ldots, n$, and effects $a_i : x_i \to L$. Let $G = \{L\}$ be the goal and $I = \{x_1 \vee \cdots \vee x_n\}$ the initial situation. The set $S_0$ of all possible initial states are the truth valuations over the $x_i$ atoms where *at least* one of these atoms is true. There are $2^n - 1$ such states. On the other hand, one can show that the set $S_0'$ of $n$ valuations in which *exactly* one of these atoms is true provides a basis for $P$; i.e., the plans that conform with these $n$ possible initial states, are exactly the plans that conform with the complete set of $2^n - 1$ possible initial states in $S_0$.

The reduction in the number of possible initial states that must be considered for computing conformant plans results from two *monotonicity properties* that we formulate using the notation $rel(s, L)$ to refer to the set of literals $L'$ that are true in the state $s$ and are relevant to the literal $L$:

$$rel(s, L) = \{L' \mid L' \in s \text{ and } L' \text{ is relevant to } L\} \ .$$

**Proposition 26** (Monotonicity 1). *Let $s$ and $s'$ be two states and let $\pi$ be an action sequence applicable in the classical problems $P/s$ and $P/s'$. Then if $\pi$ achieves a literal $L$ in $P/s'$ and $rel(s', L) \subseteq rel(s, L)$, $\pi$ achieves the literal $L$ in $P/s$.*

**Proposition 27** (Monotonicity 2). *If $S$ and $S'$ are two collections of states such that for every state $s$ in $S$ and every precondition and goal literal $L$ in $P$, there is a state $s'$ in $S'$ such that $rel(s', L) \subseteq rel(s, L)$, then if $\pi$ is a plan for $P$ that conforms with $S'$, $\pi$ is a plan for $P$ that conforms with $S$.*

From these properties, it follows that

**Proposition 28.** *$S'$ is a basis for $P$ if for every possible initial state $s$ of $P$ and every precondition and goal literal $L$ in $P$, $S'$ contains a state $s'$ such that $rel(s', L) \subseteq rel(s, L)$.*

This proposition allows us to verify the claim made in the example above that the set $S_0'$, that contains a number of states that is linear in $n$, is a basis for $P$ that has an exponential number of possible initial states. Indeed, such a problem has no precondition and a single goal literal $L$, and for every state $s$ that makes more than one atom $x_i$ true (these are the literals relevant to $L$), there is a state $s'$ in $S_0'$ that makes only one of those atoms true, and hence for which the relation $rel(s', L) \subseteq rel(s, L)$ holds.

The question that we address now is how to build a basis that complies with the condition in Proposition 28 given a covering translation $K_{T,M}(P)$. For this, let $m = \{t_1, \ldots, t_n\}$ be a merge in $M$ that covers a precondition or goal literal $L$, and let $S[t_i, L]$ denote the set of possible initial states $s$ of $P$ such that $rel(s, L) \subseteq t_i^*$; i.e., $S[t_i, L]$ contains the possible initial states of $P$ that make all the literals $L'$ that are relevant to $L$ false, except for those in the closure $t_i^*$ of $t_i$. We show first that if $I$ is in prime implicate form, $S[t_i, L]$ is a non-empty set:[7]

**Proposition 29.** *If the initial situation $I$ is in prime implicate form and $m = \{t_1, \ldots, t_n\}$ is a valid merge that covers a literal $L$ in $P$, then the set $S[t_i, L]$ of possible initial states $s$ of $P$ such that $rel(s, L) \subseteq t_i^*$ is non-empty.*

Let then $s[t_i, L]$ stand for an arbitrary state in $S[t_i, L]$. We obtain the following result:

**Theorem 30.** *Let $K_{T,M}(P)$ be a covering translation for a problem $P$ with an initial situation in PI form, and let $S'$ stand for the collection of states $s[t_i, L]$ where $L$ is a precondition or goal literal of $P$ and $t_i$ is a tag in a merge that covers $L$. Then $S'$ is a basis for $P$.*

This is an important result for three reasons. First, it tells us how to build a basis for $P$ given the tags $t_i$ in a covering translation $K_{T,M}(P)$. Second, it tells us that the size of the resulting basis is linear in the number of precondition and goal literals $L$ and tags $t_i$. And third, it makes the role of the tags $t_i$ in the covering translation $K_{T,M}(P)$ explicit, providing an intuition for why it works: each tag $t_i$ in a merge that covers a literal $L$ represents one possible initial state; namely, a state $s[t_i, L]$ that makes false all the literals $L'$ that are relevant to $L$ except those in $t_i^*$. If a plan conforms with those *critical states*, then it will conform with all the possible initial states by monotonicity (Proposition 27). It follows then in particular that:

**Theorem 31.** *If $P$ is a conformant planning problem with bounded width, then $P$ admits a basis of polynomial size.*

Namely, conformant problems $P$ with width bounded by a non-negative integer $i$ admit polynomial translations that are complete, because the plans that conform with the possibly exponential number of initial states of $P$ correspond with the plans that conform with a subset of *critical initial states* that are polynomial in number (namely, those in the polynomial basis). Thus, one complete polynomial translation for such problems is the $K_i$ translation; another one, is the $K_{S0}$ translation but with the tags associated with those critical initial states *only* rather than with all the initial states.

---

7. Recall that we are assuming throughout that the initial situation $I$ is logically consistent and that the tags $t$ are consistent with $I$.

As an illustration, for the problem $P$ above with actions $a_i$ and effects $a_i : x_i \rightarrow L$, goal $G = \{L\}$, and initial situation $I = \{x_1 \vee \cdots \vee x_n\}$, the $K_1(P)$ translation with tags $x_i$, $i = 1, \ldots, n$, and the merge $m = \{x_1, \ldots, x_n\}$ for the goal literal $L$, is a covering translation. Theorem 30 then states that a basis $S'$ for $P$ results from the collection of states $s_i$ that make each tag $x_i$ true, and all the literals that are relevant to $L$ that are not in $x_i^*$ false (i.e., all $x_k$ atoms for $k \neq i$). This is precisely the basis for $P$ that we had above that includes the states that make a single atom $x_i$ true for $i = 1, \ldots, n$: the plans that conform with this basis are then exactly the plans that conform with the whole collection of possible initial states of $P$. This basis has a size that is polynomial in $m$ though, while the number of possible initial states of $P$ is exponential in $m$.

## 8. The Planner $T_0$

The current version of the conformant planner $T_0$ is based on two instances of the general translation scheme $K_{T,M}(P)$ whose outputs are fed into the classical planner FF v2.3.[8] One instance is polynomial but not necessarily complete; the other is complete but not necessarily polynomial. For the incomplete translation, $T_0$ uses $K_1$ that is complete for problems with width no greater than 1, and as argued above, can result in solvable instances for problems of larger widths. For the complete translation, the *Kmodels* translation is used instead with a simple optimization: if the $K_1$ translation produces a single merge $m$ that covers $L$, then this merge $m$ is used for $L$ instead of the potentially more complex one determined by *Kmodels*. This is a mere optimization as the resulting translation remains complete. The other merges in *Kmodels*, that result from the models of the set of clauses $C_I(L)$ that are consistent with $I$, are computed using the SAT solver `relsat` v2.20 (Bayardo Jr. & Schrag, 1997). In the current default mode in $T_0$, which is the one used in the experiments below, the two translations $K_1$ and *Kmodels* are used in sequence: FF is called first upon the output of $K_1$ and if this fails, it is called upon the output of *Kmodels*. In the experiments below, we indicate the cases when *Kmodels* was invoked.

The translations used in $T_0$ accommodate certain simplifications and two additional actions that capture other types of deductions. The simplifications have to do with the fact that the translations considered are all *uniform* in the sense that all literals $L$ in $P$ and all rules $C \rightarrow L$ are 'conditioned' by each of the tags $t$ in $T$. From a practical point of view, however, this is not needed. The simplifications address this source of inefficiency. In particular:

- literals $KL/t$ are not created when the closure $t^*$ contains no literal relevant to $L$. In such a case, the invariance $KL/t \supset KL$ holds, and thus, every occurrence of the literal $KL/t$ in $K_{T,M}(P)$ is replaced by $KL$.

---

- support rules $a : KC/t \rightarrow KL/t$ for non-empty tags $t$ are not created when $L$ is not relevant to a literal $L'$ with a merge that contains $t$, as in such a case, the literal $KL/t$ cannot contribute to establish a precondition or goal. Similarly, cancellation rules $a : \neg K \neg C/t \rightarrow \neg K \neg L/t$ for non-empty tags $t$ are not created when $\neg L$ is not relevant to a literal $L'$ with a merge that contains $t$.

- support and cancellation rules $a : KC/t \rightarrow KL/t$ and $a : \neg K \neg C/t \rightarrow \neg K \neg L/t$ are grouped as $a : KC/t \rightarrow KL/t \wedge \neg K \neg L/t$ when for every fluent $L'$ relevant to $L$, either $L'$ or $\neg L'$ is entailed by $I$ and $t$. In such a case, there is no incomplete information about $L$ given $t$ in the initial situation, and thus the invariant $KL/t$ or $K \neg L/t$ holds, and $\neg K \neg C/t$ is equivalent to $KC/t$.

Two other types of sound deductive rules are included in the translations:

- a rule $a : KC \rightarrow KL$ is added if $a : C, \neg L \rightarrow L$ is a rule in $P$ for an action $a$, and no rule in $P$ has the form $a : C' \rightarrow \neg L$,

- rules $K \neg L_1, \ldots, K \neg L_{i-1}, K \neg L_{i+1}, \ldots, K \neg L_n \rightarrow KL_i$ for $i = 1, \ldots, n$ are added to a new unique action with no precondition, when $L_1 \vee \cdots \vee L_n$ is a static clause in $P$ (a clause in $P$ is static if true in the initial situation and provably true after any action).

These rules are versions of the *action compilation* and *static disjunctions* rules (Palacios & Geffner, 2006, 2007), and they appear to help in certain domains without hurting in others.

The version of $T_0$ reported below does not assume that the initial situation $I$ of $P$ is in prime implicate form but it rather renders it in PI form by running a version of Tison's algorithm (1967), a computation that in none of the benchmarks solved took more than 48 seconds.

The translators in $T_0$ are written in OCaml while the code for parsing the PDDL files is written in C++.

## 9. Experimental Results

We considered instances from three sources: the Conformant-FF distribution, the conformant track of the 2006 International Planning Competition (IPC5), and relevant publications (Palacios & Geffner, 2006, 2007; Cimatti et al., 2004). The instances were run on a cluster of Linux boxes at 2.33 GHz with 8GB. Each experiment had a cutoff of 2h or 2.1GB of memory. Times for $T_0$ include all the steps, in particular, computation of prime implicates, translation, and search (done by FF). We also include results from the Conformant Track of the recent 2008 International Planning Competition (IPC6).

Goals that are not sets of literals but sets of clauses are transformed in $T_0$ in a standard way: each goal clause $C : L_1 \vee \cdots \vee L_m$ is modeled by a new goal atom $G_C$, and a new action that can be executed once is added with rules $L_i \rightarrow G_C$, $i = 1, \ldots, m$.[9]

---

9. An alternative way to represent such CNF goals is by converting them into DNF first and having an action *End* map each of its non-mutex terms into a dummy goal $L_G$. This alternative encoding pays off in some cases, such as in the Adder-01 instance that does not get solved in the default CNF goal encoding (see below).

| Problem | $P$ | | | | $K_1(P)$ | | | PDDL |
|---|---|---|---|---|---|---|---|---|
| | #Acts | #Atoms | #Effects | Time | #Acts | #Atoms | #Effects | Size |
| bomb-100-100 | 10100 | 404 | 40200 | 2 | 10201 | 1595 | 50500 | 2,9 |
| square-center-96 | 4 | 196 | 760 | 35,1 | 7 | 37248 | 75054 | 3,8 |
| sortnet-09 | 46 | 68 | 109 | 8,3 | 56 | 29707 | 154913 | 5,1 |
| blocks-03 | 32 | 30 | 152 | 4 | 37 | 11370 | 35232 | 0,7 |
| dispose-16-1 | 1217 | 1479 | 2434 | 163,6 | 1218 | 133122 | 3458 | 0,3 |
| look-and-grab-8-1-1 | 352 | 358 | 2220 | 6,9 | 353 | 8708 | 118497 | 7,8 |
| sgripper-30 | 487 | 239 | 1456 | 21,5 | 860 | 1127 | 12769 | 1 |

Table 3: Translation data for selected instances: #Acts, #Atoms, and #Effects stand for the number of actions, fluents, and conditional effects. Time is the translation time in seconds rounded to the closest decimal, and PDDL Size is the size of the PDDL file in Megabytes.

Table 3 shows data concerning the translation of a group of selected instances. As it can be seen, the number of conditional effects grows considerably in all cases, and sometimes the translation may take several seconds.

Tables 4, 5, 6, 7, and 8, show the plan times and lengths obtained on a number of benchmarks by $T_0$, POND 2.2 (Bryce et al., 2006), Conformant FF (Hoffmann & Brafman, 2006), MBP (Cimatti et al., 2004) and KACMBP (Bertoli & Cimatti, 2002). These last two planners do not accept problems in the standard syntax (based on PDDL), so only a limited number of experiments were performed on them. The general picture is that $T_0$ scales up well in most domains, the exceptions being Square-Center and Cube-Center in Table 5, where KACMBP scales up better, Sortnet in Table 6, where KACMBP and MBP scale up better; and Adder in Table 6, where POND is the only planner able to solve one instance.

The problems in Table 4 are encodings from the Conformant-FF repository: Bomb-$x$-$y$ refers to the Bomb-in-the-toilet problem with $x$ packages, $y$ toilets, and clogging; Logistics-$i$-$j$-$k$ is a variation of the classical version with uncertainty about initial location of packages; Ring-$n$ is about closing and locking windows in a ring of $n$ rooms without knowing the current room; and Safe-$n$ is about opening a safe with $n$ possible combinations. All these problems have width 1. $T_0$ does clearly best on the last two domains, while in the first two domains, Conformant-FF does well too.

Table 5 reports experiments on four grid domains: Cube-Center-$n$ refers to the problem of reaching the center of a cube of size $n^3$ from a completely unknown location; Square-Center-$n$ is similar but involves square with $n^2$ possible locations; Corners-Cube-$n$ and Corners-Square-$n$ are variations of these problems where the set of possible initial locations is restricted to the Cube and Square corners respectively. MBP and KACMBP appear to be effective in these domains, although KACMBP doesn't scale up well in the corner versions. $T_0$ solves most of the problems, but in the corner versions, the quality of the plans is poor. These problems have also width 1.

Table 6 reports experiments over problems from the 2006 International Planning Competition (Bonet & Givan, 2006). The domains Coins, Comm and UTS have all width 1. The others have max width given by the number of unknown fluents in the initial situation.

| Problem | $T_0$ time | len | POND time | len | CFF time | len | MBP time | len | KACMBP time | len |
|---|---|---|---|---|---|---|---|---|---|---|
| bomb-20-1 | 0,1 | 49 | 4139 | 39 | 0 | 39 | > 2h | | 0 | 40 |
| bomb-20-5 | 0,1 | 35 | > 2h | | 0 | 35 | > 2h | | 0,2 | 40 |
| bomb-20-10 | 0,1 | 30 | > 2h | | 0 | 30 | > 2h | | 0,5 | 40 |
| bomb-20-20 | 0,1 | 20 | > 2h | | 0 | 20 | > 2h | | 2 | 40 |
| bomb-100-1 | 0,5 | 199 | – | | 56,7 | 199 | – | | 1,9 | 200 |
| bomb-100-5 | 0,7 | 195 | – | | 52,9 | 195 | – | | 4,3 | 200 |
| bomb-100-10 | 1,1 | 190 | – | | 46,8 | 190 | – | | 16,4 | 200 |
| bomb-100-60 | 4,25 | 140 | – | | 9,4 | 140 | – | | > 2h | |
| bomb-100-100 | 9,4 | 100 | – | | 1 | 100 | – | | > 2h | |
| logistics-4-3-3 | 0,1 | 35 | 56 | 40 | 0 | 37 | > 2h | | > 2.1GB | |
| logistics-2-10-10 | 1 | 84 | > 2h | | 1,6 | 83 | > 2h | | > 2.1GB | |
| logistics-3-10-10 | 1,5 | 108 | > 2h | | 4,7 | 108 | > 2h | | > 2.1GB | |
| logistics-4-10-10 | 2,5 | 125 | > 2h | | 4,4 | 121 | > 2h | | > 2.1GB | |
| ring-4 | 0,1 | 13 | 1 | 18 | 0,4 | 18 | 0 | 11 | 0 | 26 |
| ring-5 | 0,1 | 17 | 6 | 20 | 4,3 | 31 | 0,1 | 14 | 0,1 | 58 |
| ring-6 | 0,1 | 20 | 33 | 27 | 93,6 | 48 | 0,6 | 17 | 0,2 | 99 |
| ring-7 | 0,1 | 30 | 444 | 33 | 837 | 71 | 3,8 | 20 | 0,5 | 204 |
| ring-8 | 0,1 | 39 | > 2h | | > 2h | | 40 | 23 | 2 | 432 |
| ring-30 | 13,4 | 121 | – | | – | | > 2h | | > 2.1GB | |
| safe-10 | 0,1 | 10 | 0 | 10 | 0 | 10 | 0,1 | 10 | 0 | 10 |
| safe-30 | 0,1 | 30 | 2 | 30 | 1,4 | 30 | > 2h | | 0,2 | 30 |
| safe-50 | 0,4 | 50 | 9 | 50 | 29,4 | 50 | > 2h | | 0,7 | 50 |
| safe-70 | 1,12 | 70 | 41 | 70 | 109,9 | 70 | > 2h | | 2,4 | 70 |
| safe-100 | 2,5 | 100 | > 2.1GB | | 1252,4 | 100 | > 2h | | 8,6 | 100 |

Table 4: Experiments over well known benchmarks. Times reported in seconds and rounded to the closest decimal. '–' means time or memory out for smaller instances.

| Problem | $T_0$ | | POND | | CFF | | MBP | | KACMBP | |
|---|---|---|---|---|---|---|---|---|---|---|
| | time | len | time | len | time | len | time | len | time | len |
| square-center-8 | 0,2 | 21 | 2 | 41 | 70,6 | 50 | 0 | 24 | 0 | 28 |
| square-center-12 | 0,2 | 33 | 12 | 52 | > 2h | | 0 | 36 | 0 | 42 |
| square-center-16 | 0,3 | 44 | 1322 | 61 | > 2h | | 0 | 48 | 0 | 56 |
| square-center-24 | 0,8 | 69 | > 2h | | – | | 0 | 72 | 0 | 84 |
| square-center-92 | 45,3 | 273 | > 2h | | – | | 0,9 | 276 | 0,3 | 322 |
| square-center-96 | 50,2 | 285 | – | | – | | 0,9 | 288 | 0,3 | 336 |
| square-center-100 | > 2.1GB | | – | | – | | 1,1 | 300 | 0,3 | 350 |
| square-center-120 | > 2.1GB | | – | | – | | 1,9 | 360 | 0,4 | 420 |
| cube-center-5 | 0,1 | 18 | 1 | 22 | 8,2 | 45 | 0 | 28 | 0 | 25 |
| cube-center-7 | 0,1 | 27 | 2 | 43 | > 2h | | 0 | 33 | 0 | 35 |
| cube-center-9 | 0,2 | 33 | 3 | 47 | > 2h | | 0,1 | 54 | 0 | 45 |
| cube-center-11 | 0,3 | 45 | 29 | 87 | – | | 0,2 | 59 | 0 | 55 |
| cube-center-15 | 0,5 | 63 | 880 | 109 | – | | 0,2 | 69 | 0 | 75 |
| cube-center-19 | 0,8 | 81 | > 2h | | – | | 1,6 | 111 | 0,1 | 95 |
| cube-center-63 | 28,5 | 279 | > 2h | | – | | 28 | 285 | 0,5 | 315 |
| cube-center-67 | 41,6 | 297 | – | | – | | > 2.1GB | | 0,7 | 335 |
| cube-center-87 | 137,5 | 387 | – | | – | | > 2.1GB | | 1,2 | 435 |
| cube-center-91 | > 2.1GB | | – | | – | | – | | 1,2 | 455 |
| cube-center-119 | > 2.1GB | | – | | – | | – | | 2,1 | 595 |
| corners-square-12 | 0,1 | 64 | 11 | 44 | 1,7 | 82 | 0 | 36 | 0,2 | 106 |
| corners-square-16 | 0,2 | 102 | 1131 | 67 | 13,1 | 140 | 0 | 48 | 0,6 | 158 |
| corners-square-20 | 0,3 | 148 | > 2h | | 73,7 | 214 | 0,3 | 60 | 3 | 268 |
| corners-square-24 | 0,5 | 202 | > 2h | | 321 | 304 | 0,6 | 72 | 7,5 | 346 |
| corners-square-28 | 0,7 | 264 | – | | MPL | | 1,1 | 84 | 20,7 | 502 |
| corners-square-36 | 1,7 | 412 | – | | – | | 1,5 | 108 | 3308,8 | 808 |
| corners-square-40 | 2,5 | 498 | – | | – | | 7,8 | 120 | > 2h | |
| corners-square-72 | 26,1 | 1474 | – | | – | | 118,8 | 216 | > 2h | |
| corners-square-76 | 30,5 | 1632 | – | | – | | 371 | 228 | – | |
| corners-square-80 | 38,2 | 1798 | – | | – | | 649,6 | 240 | – | |
| corners-square-120 | 223,6 | 3898 | – | | – | | > 2.1GB | | | |
| corners-cube-15 | 0,8 | 147 | 907 | 105 | 134,5 | 284 | 3,7 | 69 | 174,1 | 391 |
| corners-cube-16 | 0,9 | 174 | 3168 | 115 | 439,4 | 214 | 12,5 | 72 | 270,5 | 316 |
| corners-cube-19 | 2,5 | 225 | > 2h | | 868,4 | 456 | 549,5 | 111 | 1503,1 | 488 |
| corners-cube-20 | 2,7 | 258 | > 2h | | 3975,6 | 332 | 1061,9 | 90 | 2759 | 625 |
| corners-cube-23 | 6,3 | 319 | – | | MPL | | > 2h | | 6265,9 | 899 |
| corners-cube-24 | 6,7 | 358 | – | | – | | > 2h | | > 2h | |
| corners-cube-27 | 14,6 | 429 | – | | – | | – | | > 2h | |
| corners-cube-52 | 448 | 1506 | – | | – | | – | | – | |
| corners-cube-55 | > 2.1GB | | – | | – | | – | | – | |

Table 5: Experiments over grid problems. Times reported in seconds and rounded to the closest decimal. 'MPL' for CFF means that plan exceeds maximal plan length (500 actions). '–' means time or memory out for smaller instances.

| Problem | $T_0$ time | len | POND time | len | CFF time | len | MBP time | len | KACMBP time | len |
|---|---|---|---|---|---|---|---|---|---|---|---|
| adder-01 | > 2h | | 1591 | 5 | SNH | | NR | | NR | |
| adder-02 | > 2h | | > 2h | | SNH | | NR | | NR | |
| blocks-01 | 0,1 | 5 | 0,1 | 4 | 0 | 6 | NR | | NR | |
| blocks-02 | 0,3 | 23 | 0,4 | 26 | > 2h | | NR | | NR | |
| blocks-03 | 82,6 | 80 | 126,8 | 129 | > 2h | | NR | | NR | |
| coins-10 | 0,1 | 26 | 5 | 28 | 0,1 | 38 | > 2h | | 4,2 | 106 |
| coins-12 | 0,1 | 67 | > 2h | | 0,8 | 72 | > 2h | | 3654,7 | 674 |
| coins-15 | 0,1 | 79 | > 2h | | 3 | 89 | – | | > 2h | |
| coins-16 | 0,3 | 113 | – | | 33,3 | 145 | – | | > 2h | |
| coins-17 | 0,2 | 96 | – | | 1,4 | 94 | – | | – | |
| coins-18 | 0,2 | 97 | – | | 6,2 | 118 | – | | – | |
| coins-19 | 0,2 | 105 | – | | 16,5 | 128 | – | | – | |
| coins-20 | 0,2 | 107 | – | | 20,6 | 143 | – | | – | |
| coins-21 | > 2h | | | | > 2h | | | | | |
| comm-07 | 0,1 | 54 | 0 | 47 | 0 | 47 | 0,2 | 55 | 63,6 | 53 |
| comm-08 | 0,1 | 61 | 1 | 53 | 0 | 53 | 0,2 | 71 | 1966,8 | 53 |
| comm-09 | 0,1 | 68 | 1 | 59 | 0 | 59 | 0,2 | 77 | > 2h | |
| comm-10 | 0,1 | 75 | 1 | 65 | 0 | 65 | 0,3 | 85 | > 2h | |
| comm-15 | 0,1 | 110 | 6 | 95 | 0,2 | 95 | 0,9 | 115 | – | |
| comm-16 | 0,2 | 138 | > 2h | | 0,4 | 119 | 1,6 | 151 | – | |
| comm-20 | 0,8 | 278 | > 2.1GB | | 6,4 | 239 | 50,9 | 340 | – | |
| comm-25 | 2,3 | 453 | – | | 56,1 | 389 | > 2h | | – | |
| sortnet-06 | 0,6 | 21 | 18 | 20 | SNH | | 0 | 17 | 0 | 21 |
| sortnet-07 | 2,5 | 28 | 480 | 25 | SNH | | 0 | 20 | 0 | 28 |
| sortnet-08 | 9,6 | 36 | > 2h | | SNH | | 0 | 28 | 0 | 36 |
| sortnet-09 | 76,8 | 45 | > 2h | | SNH | | 0 | 36 | 0 | 45 |
| sortnet-10 | > 2.1GB | | – | | SNH | | 0,1 | 37 | 0,1 | 55 |
| sortnet-11 | > 2.1GB | | – | | SNH | | 0,1 | 47 | 0,1 | 66 |
| uts-k-04 | 0,1 | 23 | 2 | 22 | 0,1 | 22 | 5,4 | 32 | 1,5 | 30 |
| uts-k-05 | 0,1 | 29 | 4 | 28 | 0,3 | 28 | 1247,3 | 38 | 195,4 | 42 |
| uts-k-06 | 0,2 | 35 | 10 | 34 | 0,8 | 34 | 1704,8 | 50 | > 2h | |
| uts-k-07 | 0,4 | 41 | 13 | 40 | 1,9 | 40 | > 2h | | > 2h | |
| uts-k-08 | 0,6 | 47 | 24 | 47 | 4,4 | 46 | > 2h | | – | |
| uts-k-09 | 0,9 | 53 | > 2h | | 8,6 | 52 | – | | – | |
| uts-k-10 | 1,3 | 59 | 2219 | 67 | 16,5 | 58 | – | | – | |
| uts-l-07 | 0,2 | 70 | 201 | 58 | 0,2 | 41 | 10,5 | 89 | > 2h | |
| uts-l-08 | 0,3 | 80 | 937 | 67 | 0,4 | 47 | 41,1 | 106 | > 2h | |
| uts-l-09 | 0,6 | 93 | > 2h | | 0,8 | 53 | 1176 | 137 | – | |
| uts-l-10 | 0,7 | 97 | > 2h | | 1,6 | 59 | > 2h | | – | |

Table 6: Experiments over problems from IPC5. Times reported in seconds and rounded to the closest decimal. 'SNH' for CFF means that goal syntax not handled, while 'NR' for MBP and KACMBP that these planners were not run due to lack of translations from PDDL. '–' means time or memory out for smaller instances.

| Problem | $T_0$ | | POND | | CFF | | MBP | | KACMBP | |
|---|---|---|---|---|---|---|---|---|---|---|
| | time | len | time | len | time | len | time | len | time | len |
| dispose-4-1 | 0,1 | 59 | 9 | 55 | 0,1 | 39 | > 2h | | 17,1 | 81 |
| dispose-4-2 | 0,1 | 110 | 36 | 70 | 0,2 | 56 | > 2h | | > 2h | |
| dispose-4-3 | 0,3 | 122 | 308 | 102 | 0,6 | 73 | – | | > 2h | |
| dispose-8-1 | 2,7 | 426 | > 2.1GB | | 339,1 | 227 | – | | – | |
| dispose-8-2 | 18,4 | 639 | > 2.1GB | | 2592,1 | 338 | – | | – | |
| dispose-8-3 | 197,1 | 761 | – | | > 2h | | – | | – | |
| dispose-12-1 | 78 | 1274 | – | | ME | | – | | – | |
| dispose-12-2 | 2555 | 1437 | – | | > 2.1GB | | – | | – | |
| dispose-12-3 | > 2.1GB | | – | | – | | – | | – | |
| dispose-16-1 | 382 | 1702 | – | | – | | – | | – | |
| dispose-16-2 | > 2.1GB | | | | | | | | | |
| look-and-grab-4-1-1 | 0,3 | 30 | 3098 | 16 | > 2h | | > 2h | | 0,6 | 54 |
| look-and-grab-4-1-2 | 0,5 | 4 | > 2h | | Mcl | | 0,02 | 5 | 0,0 | 6 |
| look-and-grab-4-1-3 | 0,61 | 4 | > 2h | | Mcl | | 0,01 | 5 | 0,0 | 6 |
| look-and-grab-4-2-1 | 35 | 12 | > 2.1GB | | > 2h | | > 2h | | 0,63 | 40 |
| look-and-grab-4-2-2 | 49,41 | 4 | > 2h | | Mcl | | 0,02 | 5 | 0,01 | 6 |
| look-and-grab-4-2-3 | 60,02 | 4 | > 2h | | Mcl | | 0,02 | 5 | 0,01 | 6 |
| look-and-grab-4-3-1 | > 2.1GB | | > 2.1GB | | > 2h | | > 2h | | 0,98 | 60 |
| look-and-grab-4-3-2 | 213,3 | 4 | – | | > 2h | | 0,02 | 5 | 0,02 | 6 |
| look-and-grab-4-3-3 | > 2.1GB | | – | | > 2h | | 0,02 | 5 | 0,01 | 6 |
| look-and-grab-8-1-1 | 58,2 | 242 | – | | – | | > 2h | | > 2h | |
| look-and-grab-8-1-2 | 75,3 | 90 | – | | – | | > 2h | | > 2h | |
| look-and-grab-8-1-3 | 55,89 | 58 | – | | – | | > 2h | | > 2h | |
| look-and-grab-8-2-1 | > 2h | | – | | – | | > 2h | | > 2h | |
| look-and-grab-8-2-2 | > 2h | | – | | – | | > 2h | | > 2h | |
| look-and-grab-8-2-3 | > 2h | | – | | – | | > 2h | | 1195 | 178 |
| look-and-grab-8-3-1 | > 2h | | – | | – | | > 2h | | > 2h | |
| look-and-grab-8-3-2 | > 2h | | – | | – | | > 2h | | > 2h | |
| look-and-grab-8-3-3 | > 2h | | – | | – | | > 2h | | 17,9 | 58 |

Table 7: Problems from Palacios and Geffner (2006, 2007): Times reported in seconds and rounded to the closest decimal. '–' means time or memory out for smaller instances. 'ME' and 'Mcl' mean too many edges and too many clauses respectively.

$T_0$ dominates in all these domains except in Adder where POND is the only planner able to solve an instance, and Sortnet, where MBP and KACMBP do very well, possibly due to use of the cardinality heuristic and OBDD representations. $T_0$ fails on Adder because FF gets lost in the search. Looking at this problem more closely, we found that FF could solve the (translation of the) first instance in less than a minute provided that the CNF goal for this problem is encoded in DNF as explained in footnote 9, page 646. The domains Adder, Blocks, and Sortnet in the table, along with the domain Look-and-Grab in the next table, are the only domains considered where FF run on the $K_1$ translation reports no solution after a brief search, triggering then the use of the complete *Kmodels* translation. In all the other cases where *Kmodels* was used, the $K_1$ translation had an unreachable goal fluent and there was no need to try FF on it.

| Problem | $T_0$ time | len | POND time | len | CFF time | len |
|---|---|---|---|---|---|---|
| push-to-4-1 | 0,2 | 78 | 5 | 50 | 0,3 | 46 |
| push-to-4-2 | 0,3 | 85 | 171 | 58 | 0,7 | 47 |
| push-to-4-3 | 0,6 | 87 | – | | 1,6 | 48 |
| push-to-8-1 | 81,8 | 464 | $> 2h$ | | $> 2.1GB$ | |
| push-to-8-2 | 457,9 | 423 | $> 2h$ | | $> 2.1GB$ | |
| push-to-8-3 | 1293,1 | 597 | $> 2h$ | | $> 2.1GB$ | |
| push-to-12-1 | $> 2h$ | | – | | – | |
| push-to-12-2 | $> 2h$ | | – | | – | |
| push-to-12-3 | $> 2.1GB$ | | – | | – | |
| 1-dispose-8-1 | 82,2 | 1316 | $> 2.1GB$ | | $> 2h$ | |
| 1-dispose-8-2 | $> 2.1GB$ | | $> 2.1GB$ | | $> 2h$ | |
| 1-dispose-8-3 | $> 2.1GB$ | | | | | |

Table 8: Other problems from Palacios and Geffner (2006, 2007). MBP and KACMBP were not tried on these problems as they use a different syntax. Times reported in seconds and rounded to the closest decimal. '–' means time or memory out for smaller instances.

The problems reported in Table 7 and Table 8 are variations of a family of grid problems (Palacios & Geffner, 2006, 2007). Dispose is about retrieving objects whose initial location is unknown and placing them in a trash can at a given, known location; Push-to is a variation where objects can be picked up only at two designated positions in the grid to which all objects have to be pushed to: pushing an object from a cell into a contiguous cell moves the object if it is in the cell. 1-Dispose is a variation of Dispose where the robot hand being empty is a condition for the pick up actions to work. As a result, a plan for 1-Dispose has to scan the grid, performing pick ups in every cell, followed by excursions to the trash can, and so on. The plans can get very long (a plan is reported with 1316 actions). Look-and-Grab has an action that picks up the objects that are sufficiently close if any, and after each pick-up must dump the objects it collected into the trash before continuing. For the problem P-$n$-$m$ in the table, $n$ is the grid size and $m$ is the number of objects. For Look-n-Grab, the third parameter is the radius of the action: 1 means that the hand picks up all the objects in the 8 surrounding cells, 2 that that the hand picks up all the objects in the 15 surrounding cells, and so on. The domains in Tables 7 and 8 have width 1 except 1-Dispose and Look-n-Grab. This is because, the hand being empty is a fluent that is relevant to the goal, and clauses about the location of objects are all relevant to 'hand empty'. In all these domains $T_0$ appears to do better than the other planners. The *Kmodels* translation was triggered only in the instances Look-and-Grab-$n$-$m$-$r$ for $m > 1$ (the width of these instances, as mentioned in Section 6.6, is $m$, independent of grid size).

We also report some additional data in Table 9, comparing the search that results from the use of the FF planner over the classical translations in $T_0$, to the search carried out by Conformant-FF over the original conformant problems. Conformant-FF is a conformant planner built on top of FF that searches explicitly in belief space. The table illustrates the two problems faced by belief-space planners mentioned in the introduction and the handle over them that results from the translation-based approach. The belief representation and update problem appears in the *overhead* of maintaining and evaluating the beliefs, and shows in the number of nodes that are evaluated per second: while CFF evaluates a few hundred nodes per second; FF evaluates several thousands. At the same time, the *heuristic* used in CFF in the conformant setting, appears to be less informed that the heuristic used by FF over the classical translations. In domains like Square-Center-$n$, Cube-Center-$n$, Blocks, and Look-and-Grab, FF needs orders-of-magnitude less nodes than CFF to find a plan, while the oppositive is true in Dispose-$n$-$m$ where FF evaluates many more nodes than CFF. Nonetheless, even then, due to the overhead involved in carrying the beliefs, FF manages to solve problems that CFF cannot solve. For example, the instance Dispose-8-3 is solved by $T_0$ after evaluating more than half a million nodes, but times out in CFF after evaluating less than three thousand nodes.

| Problem | CFF | | | FF in $T_0$ | | |
|---|---|---|---|---|---|---|
| | Nodes | Time | Nodes/sec | Nodes | Time | Nodes/sec |
| bomb-100-1 | 5149 | 32,9 | 156,5 | 5250 | 0,41 | 12804,9 |
| bomb-100-100 | 100 | 0,8 | 125 | 201 | 7,53 | 26,7 |
| Safe-100 | 100 | 1747,4 | 0,1 | 102 | 0 | 25500 |
| logistics-4-10-10 | 356 | 4,42 | 80,5 | 774 | 0,47 | 1646,8 |
| square-center-8 | 4634 | 59,3 | 78,1 | **46** | 0,05 | 920 |
| square-center-12 | 39000 | >5602,5 | 7 | **72** | 0,03 | 2400 |
| cube-center-5 | 2211 | 8,2 | 269,6 | **74** | 0,01 | 7400 |
| cube-center-7 | 81600 | >5602,5 | 14,6 | **105** | 0,0 | 5250 |
| blocks-01 | 46 | 0,0 | 4600 | 47 | 0 | 11750 |
| blocks-02 | 1420 | >5602,5 | 0,3 | **86** | 0,0 | 4300 |
| coins-20 | 1235 | 20,6 | 60 | 783 | 0,04 | 19575 |
| comm-25 | 517 | 56,1 | 9,2 | 1777 | 0,43 | 4132,6 |
| uts-k-10 | 58 | 16,5 | 3,5 | 62 | 0,34 | 182,4 |
| dispose-8-1 | **1107** | 339,1 | 3,3 | 11713 | 0,78 | 15016,7 |
| dispose-8-2 | **1797** | 2592,1 | 0,7 | 87030 | 14,32 | 6077,5 |
| dispose-8-3 | 2494 | >5602,5 | 0,4 | 580896 | 190,2 | 3054,1 |
| look-and-grab-4-1-1 | 4955 | >5602,5 | 0,9 | **79** | 0,1 | 790 |

Table 9: CFF over Conformant Problems vs. FF over Translations: Nodes stand for number of nodes evaluated, Time is expressed in seconds, and Nodes/sec stands for average number of nodes per second. Numbers shown in bold when either CFF or FF evaluate significantly less nodes (an order-of-magnitude reduction or more). Times preceded by '>' are time outs.

Tables 10 and 11 provide details on the results of the Conformant Track of the 2008 International Planning Competition (IPC6) (Bryce & Buffet, 2008), held almost at the time where the original version of this paper was submitted, with planner binaries submitted to the organizers a few months before. The version of $T_0$ in IPC6 was different from the version of $T_0$ used in IPC5, where it was the winning entry, and different also from the version reported in this paper. In relation, to the former, $T_0$ IPC6 was a cleaner but complete reimplementation; in relation to the latter, $T_0$ IPC6 handled problems with width greater than 1 in a different way. As explained in the previous section, the current version of $T_0$, uses $K_1$ as the basic translation regardless of the width of the problem, switching to *Kmodels* when the search over $K_1$ fails. In the version of $T_0$ at IPC6, the basic translation was a combination of $K_0$ and $K_1$; more precisely, merges for literals $L$ with width $w(L) = 1$, were generated according to $K_1$, but merges for literals $L$ with width $w(L) \neq 1$ were not generated at all. The result was that the basic translation in $T_0$ in IPC6 was lighter than the basic translation of the current version of $T_0$ but could fail on problems with width higher than 1 that the latter can solve. Retrospectively, this was not a good choice, but it didn't have much of an impact on the results. There was however a bug in the program that prevented two width-1 domains, Forest and Dispose, to be recognized as such, and thus resulted in the use of the *Kmodels* translation, that is complete for all widths, but does not scale up that well.

The other two conformant planners entered into IPC6 where CPA(H) and CPA(C); these are belief-space planners that represent beliefs as DNF formulas, and use simple belief-state heuristics for guiding the search (Tran, Nguyen, Pontelli, & Son, 2008, 2009). The belief progression in these planners is done quite effectively, by progressing each term in turn, according to the 0-approximation semantics. The potential blow up comes from the number of terms in the DNF formula encoding the initial belief state. Rather than choosing the terms of the initial belief state as the possible initial states, these planners limit the terms in the DNF formula to a collection of 'partial initial states' that do not assign any truth value to the literals that are deemed irrelevant. The resulting belief representation is complete but may still result in an exponential number of terms (Son & Tu, 2006). In order to reduce further the number of terms in this initial DNF formula, 'independent' one-of expressions are combined. For example, two independent one-of clauses $oneof(x_1, x_2)$ and $oneof(y_1, y_2)$ which would give rise to 4 possible initial states and DNF terms, are combined into the single one-of expression $oneof(x_1 \wedge y_1, x_2 \wedge y_2)$, that results into 2 possible initial

| Domain | # Instances | CpA(H) | CpA(C) | $T_0$ IPC6 |
|--------|-------------|--------|--------|------------|
| Blocks | 4 | **4** | 3 | 3 |
| Adder | 4 | 1 | **1** | 1 |
| UTS Cycle | 27 | 2 | 2 | **3** |
| Forest | 9 | 1 | 1 | **8** |
| Rao's keys | 29 | **2** | 2 | 1 |
| Dispose | 90 | **76** | 59 | 20 |

Table 10: Data from the Conformant Track of the recent IPC6 Competition: Number of problems solved by each of the conformant planners, with time out of 20 mins. In bold, entry for planner that performed best in each domain. The data is from Bryce and Buffet (2008)

states and terms. These one-of expressions are independent when they can be shown not to interact in the problem. The technique appears to be related to the notion of 'critical initial states' considered in Section 7, where it was shown that plans that conform with all critical initial states must conform also with all possible initial states. The heuristics used by CpA(H) and CpA(C) are combinations of the cardinality heuristic, that measures the number of states in a belief state, the total sum heuristic, that adds the heuristic distances to the goal from each possible state, and the number of satisfied goals, that counts the number of top goals achieved. These heuristics are all very simple, yet they work well on some benchmarks.

Tables 10 and 11 show data obtained from the IPC6 organizers from the planner logs. The first table appears in the IPC6 report (Bryce & Buffet, 2008), where the new domains Forest and Rao's keys are explained, and shows the number of problems solved by each planner, displaying in bold the planner that did best in each domain. The planner CpA(H), was declared the winner, as it was declared best in three domains (Blocks, Rao's keys, Dispose), with $T_0$ doing best in two domains (UTS Cycle and Forest), and CpA(C) doing best in one (Adder).

Table 11 shows additional details on some of the instances; in particular, the total time taken to solve the instance and the length of the plans for each of the three planners.

In terms of domain coverage, the planners do similarly on most domains, except in Forest, where $T_0$ solved most of the instances and CPA(H) solved few (8/9 vs. 1/9), and Dispose, where CPA(H) solved most of the instances and $T_0$ solved few (76/90 vs. 20/90).

In terms of time and plan quality, CpA(H) and CpA(C) appear to be slightly faster than $T_0$ on Blocks, but produce much longer plans. In Dispose, $T_0$ scales up better than CpA(H) and CpA(C) over the size of the grids, and worse on the number of objects. Indeed, only $T_0$ manages to solve the largest grid but for a single object (Dispose-10-01), and only CpA(H) and CpA(C) solve instances with more than 2 objects in the largest grids. As in most cases, plan lengths produced by $T_0$ are shorter; e.g., the plan for Dispose-04-03 contains 125 actions for $T_0$, 314 for CpA(H), and 320 for CpA(C).

Dispose is actually a domain where the cardinality heuristic does very well in the generation of plans, even if the plans tend to be rather long. As discussed above, in this domain, an agent has to scan a grid collecting a set of objects at unknown locations, and each time the action of picking up an object from a cell that may contain the object is made (except for the first time), the cardinality of the belief state is reduced. Indeed, if initially an object may be at positions $p_1$, $p_2$, ..., $p_n$, after a pick up at $p_1$, the object can be in positions $p_2, \ldots, p_n$ or in the gripper, after a pick up at $p_2$, the object can be in positions $p_3, \ldots, p_n$ or in the gripper, and so on, each pick up action decreasing the cardinality of the belief state, until becoming a singleton belief where the object must be in the gripper with certainty.

The problem with the version of $T_0$ used in IPC6 in the Dispose domain, was not only that FF explores too many states in the search, but as explained above, that it used the expensive *Kmodels* translation instead of the lighter $K_1$ translation that is complete for this domain that has width 1. With this bug fixed, $T_0$ solves 60 rather than 20 of the 90 Dispose instances, still failing on some of the larger grids with many objects, but producing much shorter plans. For example, Dispose-06-8 is solved with a plan with 470 actions, while CPA(H) and CPA(C) solve it with plans with 2881 and 3693 actions respectively. The same bug surfaced in the Forest domain, but it just prevented the solution of one instance only. Forest, Dispose, and UTS Cycle have all conformant widths equal to 1, while the other domains have all larger widths (see Table 2 for the widths of Blocks and Adder).

The second domain in IPC6 where FF got lost in the search was Adder, where indeed, $T_0$ did not solve any instance. The instance that is shown to be solved by $T_0$ in the competition report, appears to be a mistake. Similarly, the fourth instance of blocks, that is reported as solved by CPA(H), may be a mistake too; indeed, no plan for such an instance can be found in the logs, and $T_0$ reports that the goal is unreachable in the *Kmodels* translation that is complete. According to $T_0$, instance four of Rao's key is unsolvable too. On the other hand, $T_0$ failed on the larger UTS Cycle and Rao's key instances during the *translation*. In the the first, the resulting PDDL's are too large and can't be loaded into FF; in the second, the number of init clauses turns out to be quite large (above 300), giving rise to a still larger set of prime implicates (above 5000) that caused the translator to run out of memory. The second instance of Rao's keys, however, is rather small and $T_0$ didn't solve it due to a different bug. With this bug fixed, $T_0$ solves it in 0.3 seconds, producing a plan with 53 actions, which compares well with the solutions produced by CPA(H) and CPA(C) in 0.7 and 1.9 seconds, with 85 and 99 steps, respectively.

## 10. Non-Deterministic Actions

The translation schemes considered are all limited to problems with deterministic actions only. Nonetheless, as we illustrate below, these schemes can be applied to non-deterministic actions as well provided suitable transformations are included. We cover these transformations briefly as a matter of illustration only.

Consider a conformant problem $P$ with *non-deterministic* action effects $a : C \rightarrow oneof(S_1, S_2, \ldots, S_m)$, where each $S_i$ is a set (conjunction) of literals, and the transformed problem $P'$, where these effects are mapped into *deterministic* rules of the form $a : C, h_i \rightarrow S_i$, with the expression $oneof(h_1, \ldots, h_m)$ added to the initial situation of $P'$. In $P'$, the 'hidden' $h_i$ variables are used for encoding the uncertainty on the possible outcomes $S_i$ of the action $a$.

It is easy to show that the non-deterministic conformant problem $P$ and the deterministic conformant problem $P'$ are equivalent provided that only plans for $P$ and $P'$ are considered where the non-deterministic action $a$ from $P$ are executed at most once. Namely, a correspondence exists between the conformant plans for $P$ that use such actions at most once with the conformant plans for $P'$ that use the same actions at most once too. On the other hand, a conformant plan for $P'$ where these actions are done many times will not necessarily represent a conformant plan for $P$. Indeed, if $a$ non-deterministically moves an agent up or right in a square grid $n \times n$, starting in the bottom left corner, $n$ actions $a$ in a row would leave the agent at either the top left corner or the bottom right corner in $P'$, and anywhere at Manhattan distance $n$ from the origin in P. The divergence between $P$ and $P'$, however, does not arise if non-deterministic actions are executed at most once.

| Problem | Instance | CpA(H) | | CpA(C) | | $T_0$ IPC6 | |
|---|---|---|---|---|---|---|---|
| | | time | len | time | len | time | len |
| Blocks | 1 | 0 | 4 | 0 | 7 | 0,1 | 5 |
| | 2 | 0,1 | 28 | 0,1 | 35 | 0,1 | 23 |
| | 3 | 5,9 | 411 | 6,3 | 157 | 17,8 | 83 |
| | 4 | 143,9 | 257 | | | | |
| Adder | 1 | 8,5 | 3 | 8,3 | 3 | | |
| UTS Cycle | 1 | 0,8 | 3 | 0,6 | 3 | 0,1 | 3 |
| | 2 | 25,3 | 6 | 24,7 | 6 | 0,7 | 7 |
| | 3 | | | | | 5,4 | 10 |
| Forest | 1 | 3,6 | 24 | 11,6 | 18 | 0,2 | 16 |
| | 2 | | | | | 1,3 | 45 |
| | 3 | | | | | 2,2 | 78 |
| | 4 | | | | | 12,1 | 129 |
| | 5 | | | | | 14,4 | 115 |
| | 6 | | | | | 69,7 | 200 |
| | 7 | | | | | 355,1 | 256 |
| | 8 | | | | | | |
| Rao's keys | 1 | 0,1 | 28 | 0 | 29 | 0 | 16 |
| | 2 | 0,7 | 85 | 1,9 | 99 | | |
| Dispose | 4,1 | 0,3 | 80 | 0,4 | 88 | 0,1 | 77 |
| | 4,2 | 0,7 | 197 | 0,9 | 206 | 3,6 | 110 |
| | 4,3 | 1,3 | 314 | 1,8 | 320 | 528,3 | 125 |
| | 4,4 | 2 | 431 | 2,8 | 434 | | |
| | 6,1 | 4,7 | 270 | 4,5 | 187 | 0,9 | 204 |
| | 6,2 | 10,4 | 643 | 42,2 | 735 | 217,7 | 329 |
| | 6,3 | 17,7 | 1016 | 97,9 | 1228 | | |
| | 6,4 | 27,6 | 1389 | 172,5 | 1721 | | |
| | 8,1 | 40,1 | 753 | 40,3 | 518 | 7,4 | 326 |
| | 8,2 | 86,7 | 1851 | 524,6 | 1962 | | |
| | 8,3 | 86,7 | 1851 | | | | |
| | 10,1 | | | | | 45 | 683 |
| | 10,2 | | | | | | |

Table 11: Running time and plan length from IPC6 logs. Time in seconds. Blanks stand for time or memory out. Only 13 of the 90 Dispose-*n-m* instances shown, At IPC6, size *n* of grid ranged from 2 to 10, while number *m* of objects, from 1 to 10. $T_0$ scales up best on *n* and worst on *m*.

Building on this idea, a non-deterministic conformant planner can be obtained from a deterministic conformant planner in the following way. For the non-deterministic problem $P$, let $P_1$ be the problem $P'$ above, with the additional constraint that the actions $a$ in $P_1$ arising from the non-deterministic actions in $P$ can be executed at most once. This is easily achieved by adding a precondition $enabled(a)$ to $a$ that is true initially and that $a$ sets to false. Let then $P_2$ represent the deterministic conformant problem where each non-deterministic action $a$ in $P$ is mapped into 2 deterministic actions, each executable only once, and each having its own 'hidden fluents' $h_1, \ldots, h_m$ with the $oneof(h_1, \ldots, h_m)$ expression in the initial situation. Similarly, let $P_i$ be the deterministic problem that results from encoding each non-deterministic action in $P$ with $i$ deterministic 'copies'.

From this encoding, a simple iterative conformant planner for non-deterministic problems $P$ can be defined in terms of a conformant planner for deterministic problems by invoking the latter upon $P_1$, $P_2$, $P_3$, and so on, until a solution is reported. The reported solution uses each copy of a 'non-deterministic action' at most once, and thus encodes a solution to the original problem.

We have implemented this strategy on top of $T_0$ with an additional refinement that takes advantage of the nature of the $K_{T,M}$ translation, where assumptions about the initial situation are maintained explicitly in tags. Basically, 'non-deterministic' actions $a$ in $P_i$ are allowed to be executed more than once provided that all the literals $KL/h_i$ that depend on a particular outcome of these actions ($S_i$) are erased. This is implemented by means of an additional $reset(a)$ action in $P_i$ whose unconditional effect is $enabled(a)$ (i.e., the action $a$ can then be done again) and whose conditional effects are $\neg KL \to \neg KL/h_i$ and $KL \to KL/h_i$ for $i = 1, \ldots, m$. Namely, literals $KL/h_i$ where the truth of $L$ depends on a particular non-deterministic outcome ($S_i$) are erased, except when $L$ is true with no assumptions; i.e. when $KL$ is true. Then non-deterministic actions $a$ can be executed more than once in a plan provided that each occurrence of $a$, except for the first one, is preceded by a $reset(a)$ action.

Table 12 compares the resulting non-deterministic planner with MBP and KACMBP on a number of non-deterministic problems considered in the MBP and KACMBP papers. We have just added an additional domain, Slippery Gripper (sgripper), that is similar to classical Gripper where a number of balls have to be moved from room $A$ to $B$, except that the robot cannot move from $A$ to $B$ directly, but has a non-deterministic move action $move(A, C, D)$ that moves the robot from $A$ to either $C$ or $D$. A typical plan for moving two balls from $A$ to $B$ is to pick them at $A$, move to $C$ or $D$, move from $C$ to $B$, and from $D$ to $B$, finally dropping the balls at $B$.

For the deterministic conformant planner ($T_0$) used in the non-deterministic setting we added the following modification: merges are not introduced only for precondition and goal literals but for all literals. The reason is that in this setting it pays to remove the uncertainty of all literals when the reset mechanism is used. Indeed, provided with this simple change and the reset mechanism, in none of the problems we had to move beyond $P_1$ (a single copy of each non-deterministic action) even if in all the domains non-deterministic actions are required many times in the plans (e.g., if there are more than 2 balls in room A).

| Problem | $T_0$ time | len | MBP time | len | KACMBP time | len |
|---|---|---|---|---|---|---|
| sgripper-10 | 1,4 | 48 | > 2h | | 0,6 | 68 |
| sgripper-20 | 16,7 | 93 | > 2h | | 5,4 | 148 |
| sgripper-30 | 90 | 138 | – | | 23,3 | 228 |
| btuc-100 | 2,9 | 200 | > 2h | | 2 | 200 |
| btuc-150 | 9,2 | 300 | > 2h | | 7,9 | 300 |
| btuc-200 | 23 | 400 | – | | 16,9 | 400 |
| btuc-250 | 44,6 | 500 | – | | 33,2 | 500 |
| btuc-300 | 82 | 600 | – | | 62,1 | 600 |
| bmtuc-10-10 | 0,1 | 20 | 65,9 | 29 | 0,2 | 20 |
| bmtuc-20-10 | 0,1 | 40 | > 2h | | 0,6 | 40 |
| bmtuc-20-20 | 0,3 | 40 | > 2h | | 2,2 | 40 |
| bmtuc-50-10 | 0,9 | 100 | – | | 3,6 | 100 |
| bmtuc-50-50 | 3,3 | 100 | – | | 2722,4 | 100 |
| bmtuc-100-10 | 4,9 | 200 | – | | 25,1 | 200 |
| bmtuc-100-50 | 14,9 | 200 | – | | > 2h | |
| bmtuc-100-100 | 30,2 | 200 | – | | > 2h | |
| nondet-ring-5 | 18,3 | 19 | 0 | 18 | 0,1 | 32 |
| nondet-ring-10 | > 2h | | 2,1 | 38 | 0,5 | 112 |
| nondet-ring-15 | > 2h | | 1298,9 | 58 | 2,4 | 242 |
| nondet-ring-20 | – | | > 2h | | 7,3 | 422 |
| nondet-ring-50 | – | | – | | 603,1 | 2552 |
| nondet-ring-1key-5 | > 2h | | 0,1 | 33 | 0,2 | 42 |
| nondet-ring-1key-10 | > 2.1GB | | 11,2 | 122 | 4 | 197 |
| nondet-ring-1key-15 | – | | 5164,4 | 87 | 33,7 | 375 |
| nondet-ring-1key-20 | – | | > 2.1GB | | 246,5 | 1104 |
| nondet-ring-1key-25 | – | | – | | 1417,5 | 2043 |
| nondet-ring-1key-30 | – | | – | | > 2h | |

Table 12: Non-deterministic problems. All problems except sgripper are from MBP and KACMBP. These problems were modified to render a simple translation into PDDL; in particular, complex preconditions were moved in as conditions. Times reported in seconds and rounded to the closest decimal. '–' means time or memory out for smaller instances.

As it can be seen from the table, $T_0$ does better than MBP on these collection of non-deterministic domains, although not as well as KACMBP, in particular, in the NonDet-Ring and Non-Det-Ring-1Key domains. In any case, the results obtained with $T_0$ on these domains are quite meaningful. In all cases where $T_0$ failed to solved a problem, the reason was that the classical planner (FF) got lost in the search for plans, something that may improve with further advances in classical planning technology.

## 11. Related Work

Most recent conformant planners such as CFF, POND, and MBP cast conformant planning as an heuristic search problem in belief space (Bonet & Geffner, 2000). Compact belief representations and informed heuristic functions, however, are critical for making these approach work. As an effective belief representation, these planners use SAT and OBDDs techniques that while intractable in the worst case often exhibit good behavior on average. As heuristics, on the other hand, they use fixed cardinality heuristics that count the number of states that are possible for a given belief state (a tractable operation on OBDD representations) or heuristics obtained from a relaxed planning graph suitably extended to take uncertain information into account. These heuristics appear to work well in some domains but not in others. From this perspective, the translation-based approach provides a handle on the two problems: belief states in $P$ become plain states in the translation $K_{T,M}(P)$, that is then solved using classical heuristics. We have also established the conditions under which this belief representation is compact and complete.

A sound but incomplete approach to planning with incomplete information is advanced by Petrick and Bacchus (2002) that represent belief states as formulas. In order to make belief updates efficient though, several approximations are introduced, and in particular, while existing disjunctions can be carried from one belief to the next, no new disjunctions are added. This imposes a limitation on the type of problems that can be handled. The two other limitations of this approach are that domains must be crafted by hand, and that no control information is derived from the domains so that the search for plans is blind. Our approach can be understood as providing a solution to these two problems too: on the one hand, the move to the 'knowledge-level' is done automatically, on the other, the problem lifted to the knowledge-level is solved by classical planners able to search with control information derived automatically from the new representation.

A third thread of work related to our approach arises from the so-called 0-approximation semantics (Baral & Son, 1997). In the 0-approximation semantics, belief states $b$ are represented not by sets of states but by a single 3-valued state where fluents can be true, false, or unknown. In Proposition 3 above, a correspondence was established between the plans for $P$ that are conformant according to the 0-approximation semantics and the classical plans for the translation $K_0(P)$, which in turns is an instance of the more general translation $K_i(P)$ that is complete for problems with width $i = 0$. The semantics of the translation $K_0$ is thus related to the 0-approximation semantics, yet the $K_0$ translation delivers something more: a computational method for obtaining conformant plans that comply with the 0-approximation semantics using a classical planner.

The 0-approximation and the basic $K_0$ translation are too weak for dealing with the existing benchmarks. The translations $K_i$ extend $K_0$ for problems of higher width by replacing the set of fluents $KL$ by fluents $KL/t$ where the tags $t$ encode assumptions about the initial situation. The extensions of the 0-approximation semantics in the context of conformant planning have taken a different form: switching from a single 3-valued state for representing beliefs to *sets* of 3-valued states, each 3-valued state progressed efficiently and independently of the others (Son, Tu, Gelfond, & Morales, 2005). The initial set of 3-valued states is obtained by forcing states to assign a boolean truth-value (true or false) to a number of fluents. Crucial for this approach to work is the number of such fluents;

belief representation and update are exponential in it. The conditions that ensure the completeness of this extension of the 0-approximation semantics can be expressed in terms of a relevance analysis similar to the one underlying our analysis of width (Son & Tu, 2006): the fluents that must be set to true or false in each initial 3-valued state are those appearing in a clause in $C_I(L)$ for a precondition or goal literal $L$. In particular, if in the initial situation there are $n$ tautologies $p_i \vee \neg p_i$, each relevant to a precondition or goal literal $L$, then the number of initial 3-valued states required for completeness is exponential in $n$, as each has to make each fluent $p_i$ true or false. The difference with our approach can be seen when each of the tautologies $p_i \vee \neg p_i$ is relevant to a *unique* precondition or goal literal $L_i$. In such a case, the number of 3-valued or 'partial' states required for completeness remains exponential in $n$, while the resulting problem has width 1 and thus can be solved with the $K_1$ translation that involves tags with a single literal. In other words, while the tags used in our translation scheme encode the *local contexts* required by the different literals in the problem, the initial 3-valued states (Son & Tu, 2006) encode their possible combinations in the form of *global contexts*. These global contexts correspond to the consistent combinations of such local contexts, which may thus be exponential in number even if the problem has bounded width. The planners CPA(H) and CPA(C), discussed above in the context of the Conformant Track of the recent 2008 Int. Planning Competition (IPC6), build on this approach, but reduce the number of partial initial states required using a technique that can replace many one-of expressions by a single one (Tran et al., 2008, 2009); a simplification related to the notion of 'critical' initial states discussed in Section 7.

Another difference with the 3-valued approach (Son et al., 2005; Son & Tu, 2006), is that the translation approach not only addresses the representation of beliefs but also the computation of conformant plans: once a conformant problem $P$ is translated into a problem $K_{T,M}(P)$, it can be solved by a classical planner. The approaches that have been defined on top of the 0-approximation semantics, like the knowledge-level approach to planning with incomplete information by Petrick and Bacchus (2002), need a way to guide the search for plans in the simplified belief space. While the search by Petrick and Bacchus (2002) is blind (iterative deepening), the search by Son et al. (2005), Son and Tu (2006) is guided by a combination of simple heuristics such as cardinality or subgoal counting.

## 12. Summary

While few practical problems are purely conformant, the ability to find conformant plans is needed in contingent settings where conformant situations are an special case. In this paper, we have introduced a new approach to conformant planning where conformant problems $P$ are converted into classical planning problems $K_{T,M}(P)$ that are then solved by a classical planner. We have also studied the conditions under which this general translation is sound and complete. The translation depends on two parameters: a set of tags, referring to local contexts in the initial situation, and a set of merges that stand for valid disjunctions of tags. We have seen how different translations, such as $K_{S0}$ and $Kmodels$, can be obtained from suitable choices of tags and merges, and have introduced a measure of complexity in conformant planning called *conformant width*, and a translation scheme $K_i$ that is polynomial for a fixed $i$ and complete for problems with width bounded by $i$. We have also shown that most conformant benchmarks have width 1, have developed a conformant planner $T_0$ based on these translations, and have shown that this planner exhibits a good performance in comparison with existing conformant planners. Recently, we have explored the use of these ideas in the more general setting of contingent planning (Albore, Palacios, & Geffner, 2009).

## Acknowledgments

We thank Alex Albore for help with the syntax of MBP and KACMBP, and Pierre Marquis for kindly answering a question about the complexity of a deductive task. We also thank the anonymous reviewers for useful comments. H. Geffner is partially supported by grant TIN2006-15387-C03-03.

## Appendix A. Proofs

$P$ below stands for a conformant planning problem $P = \langle F, I, O, G \rangle$ and $K_{T,M}(P) = \langle F', I', O', G' \rangle$ for its translation. Propositions and theorems in the body of the paper appear in the appendix with the same numbers; while new lemmas and propositions have numbers preceded by the letters A and B (for Appendix A and B). The conformant problem $P$ and the classical problems $P/s$ and $K_{T,M}(P)$ that arise from $P$ are all assumed to be *consistent*. Consistency issues are important, and they are addressed in more detail in the second part of this appendix where it is shown that if $P$ is consistent, $K_{T,M}(P)$ is consistent too (Appendix B). For a consistent *classical* problem $P'$, the standard progression lemma applies; namely, a literal $L$ is achieved by an applicable action sequence $\pi_{+1} = \pi, a$, where $\pi$ is an action sequence and $a$ is an action iff A) $\pi$ achieves $C$ for a rule $a : C \rightarrow L$ in $P'$, or B) $\pi$ achieves $L$ and the negation $\neg L'$ of a literal $L'$ in the body $C'$ of each rule in $P'$ of the form $a : C' \rightarrow \neg L$ (see Theorem B.2 below).

**Lemma A.1.** *Let $\pi$ be an action sequence applicable in both $P$ and $K_0(P)$. Then if $\pi$ achieves $KL$ in $K_0(P)$, $\pi$ achieves $L$ in $P$.*

*Proof.* By induction on the length of $\pi$. If $\pi$ is empty and $\pi$ achieves $KL$ in $K_0(P)$, then $KL$ must be in $I'$, and hence $L$ must be in $I$, so that $\pi$ achieves $L$ in $P$.

Likewise, if $\pi_{+1} = \pi, a$ achieves $KL$ in $K_0(P)$ then *A)* there is rule $a : KC \rightarrow KL$ in $K_0(P)$, such that $\pi$ achieves $KC$ in $K_0(P)$; or *B)* $\pi$ achieves $KL$ in $K_0(P)$ and for each rule $a : \neg K \neg C' \rightarrow \neg KL$ in $K_0(P)$, $\pi$ achieves $K \neg L'$ in $K_0(P)$ for some $L'$ in $C'$.

If *A)* is true, then $P$ must contain a rule $a : C \rightarrow L$, and by inductive hypothesis, $\pi$ must achieve $C$ in $P$, and therefore, $\pi_{+1} = \pi, a$ must achieve $L$ in $P$. If *B)* is true, by inductive hypothesis, $\pi$ must achieve $L$ in $P$ along with $\neg L'$ for some literal $L'$ in the body $C'$ of each rule $a : C' \rightarrow \neg L$, and thus $\pi_{+1} = \pi, a$ must achieve $L$ in $P$ too. $\square$

**Lemma A.2.** *If an action sequence $\pi$ is applicable in $K_0(P)$, then $\pi$ is applicable in $P$.*

*Proof.* If $\pi$ is empty, this is trivial. Likewise, if $\pi_{+1} = \pi, a$ is applicable in $K_0(P)$, $\pi$ is applicable in $K_0(P)$, and thus by inductive hypothesis, $\pi$ is applicable in $P$. Also since, $\pi, a$ is applicable in $K_0(P)$, $\pi$ must achieve the literals $KL$ in $K_0(P)$ for each precondition $L$ of $a$, but then from Lemma A.1, $\pi$ must achieve the literals $L$ for the same preconditions in $P$, and thus, the sequence $\pi_{+1} = \pi, a$ is applicable in $P$. $\square$

**Proposition 2** *If $\pi$ is a classical plan for $K_0(P)$, then $\pi$ is a conformant plan for $P$.*

*Proof.* Direct from Lemma A.2 once we consider a problem $P'$ similar to $P$ but with a new dummy action $a_G$ whose preconditions are the goals $G$ of $P$. Then if $\pi$ is a plan for $K_0(P)$, $\pi, a_G$ is applicable in $K_0(P')$, and by Lemma A.2, $\pi, a_G$ is applicable in $P'$, which implies that $\pi$ is applicable in $P$ and achieves $G$, and thus, that $\pi$ is a plan for $P$. $\qquad\square$

**Proposition 3** *An action sequence $\pi$ is a classical plan for $K_0(P)$ iff $\pi$ is a conformant plan for $P$ according to the 0-approximation semantics.*

*Proof.* Let us say that an action sequence $\pi = a_0, \ldots, a_n$ is 0-applicable in $P$ and 0-achieves a literal $L$ in $P$ if the belief sequence $b_0, \ldots, b_{n+1}$ generated according to the 0-approximation semantics is such that the preconditions of the actions $a_i$ in $\pi$ are true in $b_i$, and the goals are true in $b_{n+1}$ respectively. From the definition of the 0-approximation semantics (and the consistency of $P$), an applicable action sequence $\pi$ thus 0-achieves a literal $L$ in $P$ iff $\pi$ is empty and $L \in I$, or $\pi = \pi', a$ and A) $a : C \to L$ is an effect of $P$ and $\pi'$ 0-achieves each literal $L'$ in $C$, or B) $\pi'$ 0-achieves $L$ and for all effects $a : C' \to \neg L$ in $P$, $\pi'$ 0-achieves $\neg L'$ for some $L' \in C'$. These, however, are the conditions under which $\pi$ achieves the literal $KL$ in $K_0(P)$ once 'a sequence 0-achieving a literal $L$ in $P$' is replaced by 'a sequence achieving the literal $KL$ in $K_0(P)$'. Thus, an action sequence $\pi$ that is applicable in $K_0(P)$ and 0-applicable in $P$ achieves a literal $KL$ in $K_0(P)$ iff $\pi$ 0-achieves the literal $L$ in $P$, while $\pi$ is applicable to $K_0(P)$ iff it is 0-applicable to $P$, with the last part following from the first using induction on the plan length. $\qquad\square$

**Definition A.3.** For an action $a$ in $P$, define $a^*$ to be the action sequence where $a$ is followed by all merges in $K_{T,M}(P)$ in arbitrary order. Similarly, if $\pi = a_0, \ldots, a_i$ is an action sequence in $P$, define $\pi^*$ to be the action sequence $\pi^* = a_0^*, \ldots, a_n^*$ in $K_{T,M}(P)$.

**Lemma A.4.** *Let $\pi$ be an action sequence such that $\pi$ is applicable in $P$ and $\pi^*$ is applicable in a valid translation $K_{T,M}(P)$. If $\pi^*$ achieves $KL/t$ in $K_{T,M}(P)$, then $\pi$ achieves $L$ in $P/s$ for all possible initial states $s$ that satisfy $t$.*

*Proof.* For an empty $\pi$, if $\pi^*$ achieves $KL/t$, from the definition of $K_{T,M}(P)$ and since $I \models t \supset L$, $L$ must be in any such $s$, and thus $\pi$ must achieve $L$ in $P/s$.

Likewise, if $\pi_{+1} = \pi, a$ and $t$ is *not* the empty tag, $\pi_{+1}^* = \pi^*, a^*$ achieves $KL/t$ in $K_{T,M}(P)$ iff A) $\pi^*$ achieves $KC/t$ in $K_{T,M}(P)$ for a rule $a : KC/t \to KL/t$ in $K_{T,M}(P)$, or B) $\pi^*$ achieves $KL/t$, and for any rule $a : \neg K\neg C'/t \to \neg KL/t$, $\pi^*$ achieves $K\neg L'/t$ in $K_{T,M}(P)$ for some $L'$ in $C'$ (merge actions do not delete positive literals $KL/t$).

If A, by inductive hypothesis, $\pi$ achieves $C$ in $P/s$ for each possible initial state $s$ that satisfies $t$, and hence $\pi_{+1} = \pi, a$ achieves $L$ in $P/s$ from the rule $a : C \to L$ that must be in $P$. If B, by inductive hypothesis, $\pi$ achieves $L$ and $\neg L'$ in $P/s$, for some $L'$ in the body of each rule $a : C' \to \neg L$ in $P$, and thus $\pi_{+1} = \pi, a$ achieves $L$ in $P/s$.

For the empty tag $t = \emptyset$, a third case must be considered: a merge action $\bigwedge_{t' \in m} KL/t' \to KL$ in $a^*$ may be the cause for the action sequence $\pi_{+1}^* = \pi^*, a^*$ achieving $KL$ in $K_{T,M}(P)$. In such a case, the sequence $\pi^*, a$, and hence $\pi^*, a^*$, must achieve $KL/t'$ for each (non-empty) $t'$ in $m$ in $K_{T,M}(P)$, and hence from the inductive hypothesis and the two cases above, the sequence $\pi, a$ must achieve $L$ in $P/s$ for each possible initial state $s$ that satisfies any such $t'$. Yet, since the merge $m$ is valid, all possible initial states $s$ must satisfy one such $t'$, and thus $\pi$ must achieve $L$ in $P/s$ for all possible initial states $s$, that are the initial states that satisfy $t = \emptyset$. □

**Lemma A.5.** *If $\pi^*$ is applicable in a valid translation $K_{T,M}(P)$, then $\pi$ is applicable in $P$.*

*Proof.* If $\pi$ is empty, this is direct. For $\pi_{+1} = \pi, a$, if $\pi_{+1}^* = \pi^*$, $a^*$ is applicable in $K_{T,M}(P)$, then $\pi^*$ is applicable in $K_{T,M}(P)$, achieving $KL$ for each precondition $L$ of $a$, and hence from the inductive hypothesis, $\pi$ is applicable in $P$, and from Lemma A.4, $\pi$ must achieve $L$ for each precondition $L$ of $a$, and thus $\pi_{+1} = \pi, a$ is applicable in $P$. □

**Theorem 7** *The translation $K_{T,M}(P)$ is sound provided that all merges in $M$ are valid and all tags in $T$ are consistent.*

*Proof.* Consider the problem $P'$ that is similar to $P$ but with a new dummy action $a_G$ whose preconditions are the goals $G$ of $P$. We have then that $\pi^*$ is a plan for $K_{T,M}(P)$ iff $\pi_1^*, a_G^*$ is applicable in $K_{T,M}(P')$, which from Lemma A.5 implies that $\pi, a_G$ is applicable in $P'$, which means that $\pi$ is a plan for $P$. □

**Lemma A.6.** *Let $\pi$ be an action sequence such that $\pi$ is applicable in $P$ and $\pi^*$ is applicable in $K_{S0}(P)$. If $\pi$ achieves $L$ in $P/s$ for some possible initial state $s$, $\pi^*$ achieves $KL/s$ in $K_{S0}(P)$.*

*Proof.* If $\pi$ is empty and $\pi$ achieves $L$ in $P/s$, then $L \in s$, and since $I \models s \supset L$, $KL/s$ must be in $I'$ and thus $\pi^*$ achieves $KL/s$ in $K_{S0}(P)$.

Likewise, if $\pi_{+1} = \pi, a$ achieves $L$ in $P/s$ then A) there is rule $a : C \to L$ such that $\pi$ achieves $C$ in $P/s$; or B) $\pi$ achieves $L$ and for any rule $a : C' \to \neg L$, $\pi$ achieves $\neg L'$ in $K_{S0}(P)$ for some $L' \in C'$.

If A), by inductive hypothesis, $\pi^*$ achieves $KC/s$ in $K_{S0}(P)$ and, from rule $a : KC/s \to KL/s$, $\pi^*, a$ must achieve $KL/s$, and thus, $\pi_{+1}^* = \pi^*, a^*$ achieves $KL/s$ (merges in $a^*$ do not delete positive literals $KL/t$).

If B), by inductive hypothesis, $\pi^*$ achieves $KL/s$ and $K\neg L'/s$ in $K_{S0}(P)$ for some $L'$ in the body of each rule $a : C' \to \neg L$ in $P$, and therefore $\pi^*, a$ achieves $KL/s$, and so does $\pi_{+1}^* = \pi^*, a^*$. □

**Lemma A.7.** *If $\pi$ is applicable in $P$, $\pi^*$ is applicable in $K_{S0}(P)$.*

*Proof.* If $\pi$ is empty, this is trivial. If $\pi_{+1} = \pi, a$ is applicable in $P$, then $\pi$ must be applicable in $P$ and must achieve each precondition $L$ of $a$ in $P/s$ for every possible initial state $s$, $s \in S_0$. From the inductive hypothesis, $\pi^*$ must then be applicable in $K_{S0}(P)$, and from Lemma A.6, it must achieve the literals $KL/s$ for all $s \in S_0$, and then, the last merge action with effect $\bigwedge_{s \in S_0} KL/s \to KL$ in $\pi^*$ must achieve $KL$, and so does $\pi^*$, and therefore, $\pi^*, a^*$ is applicable in $K_{S0}(P)$. □

**Theorem 9** *If $\pi$ is a conformant plan for $P$, then there is a classical plan $\pi'$ for $K_{S0}(P)$ such that $\pi$ is the result of dropping the merge actions from $\pi'$.*

*Proof.* Direct from Lemma A.7 if we consider a problem $P'$ similar to $P$ but with a new action $a_G$ whose preconditions are the goals $G$ of $P$. If $\pi$ is a plan for $P$, the sequence $\pi, a_G$ is applicable in $P'$, and from Lemma A.7, $\pi^*, a_G^*$ is applicable in $K_{S0}(P')$, and thus $\pi^*$ is a plan for $K_{S0}(P)$. $\qquad\square$

**Definition A.8.** $rel(s, L)$ stands for the set of literals $L'$ in $s$ that are relevant to $L$ in $P$:

$$rel(s, L) = \{L' \mid L' \in s \text{ and } L' \text{ is relevant to } L\} \ .$$

**Definition A.9.** $t^*$ stands for the deductive closure of $t$ under $I$:

$$t^* = \{ \ L \mid I, t \models L\} \ .$$

**Theorem A.10.** *Let $m = \{t_1, \dots, t_n\}$ be a covering merge for a literal $L$ in a valid translation $K_{T,M}(P)$ for a problem $P$ whose initial situation is in prime implicate form. Then for each tag $t_i$ in $m$ there must be a possible initial state $s$ of $P$ such that $rel(s, L) \subseteq t_i^*$.*

*Proof.* Assume otherwise that each state $s$ satisfying $I$ makes true a literal $L_s$ relevant to $L$ such that $L_s \notin t_i^*$. If we then take $c$ to be the disjunction of such literals $L_s$ over all the states $s$ that satisfy $I$, we obtain that $I$ entails $c$, which since $I$ is in prime implicate form, means that $c$ contains a tautology $c'$ or is subsumed by a clause $c''$ in $I$. But, in either case, this is a contradiction, as all the literals in $c'$ or $c''$ are relevant to $L$, where $t_i^*$, where $t_i$ is part of the covering merge $m$, must contain a literal in either $c'$ or $c''$, and hence in $c$. $\quad\square$

**Lemma A.11.** *Let $\pi$ be an action sequence such that $\pi$ is applicable in $P$ and $\pi^*$ is applicable in a covering translation $K_{T,M}(P)$. Then, if $\pi$ achieves $L$ in $P/s$ for some possible initial state $s$ and there is a tag $t$ in $T$ such that $rel(s, L) \subseteq t^*$, $\pi^*$ achieves $KL/t$ in $K_{T,M}(P)$.*

*Proof.* If $\pi$ is empty and $\pi$ achieves $L$ in $P/s$, then $L$ is in $s$ and thus, in $rel(s, L)$. Since $rel(s, L) \subseteq t^*$, then $L \in t^*$, and thus $KL/t$ is in the initial situation $I'$ of $K_{T,M}(P)$, and $\pi^*$ achieves $KL/t$ in $K_{T,M}(P)$. Likewise, if $\pi_{+1} = \pi, a$ achieves $L$ in $P/s$, then A) there is a rule $a : C \to L$ in $P$ such that $\pi$ achieves $C$ in $P/s$, or B) $\pi$ achieves $L$ in $P/s$ and for each rule $a : C' \to \neg L$, $\pi$ achieves $\neg L'$ in $P/s$ for some $L'$ in $C'$. If A, by inductive hypothesis, $\pi^*$ achieves $KC/t$, and from the support rule $a : KC/t \to KL/t$ in $K_{T,M}(P)$, $\pi^*, a$ must achieve $KL/t$ in $K_{T,M}(P)$, and so must $\pi_{+1}^* = \pi^*, a^*$, as the merges in $a^*$ cannot delete a positive literal $KL/t$. If B, by inductive hypothesis, $\pi^*$ achieves $KL/t$, and for each cancellation rule $a : \neg K \neg C'/t \to \neg KL/t$ arising from the rule $a : C' \to \neg L$ in $P$, $\pi^*$ must achieve $K \neg L'/t$ for some literal $L' \in C'$. This means that $\pi^*, a$, and therefore, $\pi_{+1}^* = \pi^*, a^*$, must achieve $KL/t$. $\quad\square$

**Lemma A.12.** *Let $K_{T,M}(P)$ be a covering translation of $P$. Then if $\pi$ is applicable in $P$, $\pi^*$ is applicable in $K_{T,M}(P)$.*

*Proof.* If $\pi$ is empty, this is direct. Else, if $\pi_{+1} = \pi, a$ is applicable in $P$, then $\pi$ must be applicable in $P$ where it must achieve each literal $L$ in $Pre(a)$, and therefore, by inductive hypothesis $\pi^*$ must be applicable in $K_{T,M}(P)$. Then, let $m = \{t_1, \dots, t_n\}$ be a covering merge for $L \in Pre(a)$ in $K_{T,M}(P)$. From Theorem A.10, for each $t_i \in m$ there must be a possible initial state $s$ such that $rel(s, L) \subseteq t_i^*$, and then from Lemma A.11, $\pi$ achieving $L$ in $P/s$ implies $\pi^*$ achieving $KL/t_i$ in $K_{T,M}(P)$. Since this is true for all $t_i \in m$ and $\pi$ achieves $L \in Pre(a)$ in $P/s$ for all possible initial states $s$, then it follows that $\pi^*$ achieves $KL/t_i$ for all $t_i \in m$ in $K_{T,M}(P)$, and therefore that $\pi^*$ achieves $KL$ in $K_{T,M}(P)$ as $\pi^*$ ends with a sequence of merges that include the action merge $a_{m,L}$ with effect $\bigwedge_{t_i \in m} KL/t_i \rightarrow KL$. As a result, $\pi_{+1}^* = \pi^*, a^*$ is applicable in $K_{T,M}(P)$. $\qquad\square$

**Theorem 15** *Covering translations $K_{T,M}(P)$ are complete; i.e., if $\pi$ is a conformant plan for $P$, then there is a classical plan $\pi'$ for $K_{T,M}(P)$ such that $\pi$ is $\pi'$ with the merge actions removed.*

*Proof.* The theorem follows trivially from Lemma A.12 by having a problem $P'$ that is like $P$ but with an additional, dummy action $a_G$ such that the goals $G$ of $P$ are the preconditions of $a_G$. The action sequence $\pi$ is a plan for $P$ iff the action sequence $\pi, a_G$ is applicable in $P'$, which due to Lemma A.12 implies that the action sequence $\pi^*, a_G^*$ is applicable in $K_{T,M}(P')$ which in turn is true iff the action sequence $\pi^*$ is a plan for $K_{T,M}(P)$. The sequence $\pi$, in turn, is the sequence $\pi^*$ with all the merge actions removed. $\qquad\square$

**Theorem 17** *The translation Kmodels($P$) is sound and complete.*

*Proof.* Direct from the merges $m$ generated by *Kmodels* for each precondition and goal literals $L$. Clearly these merges are all valid, their tags are consistent with $I$, and they cover $L$ (the models of $C_I(L)$ all satisfy $C_I(L)$). Thus the result follows from Theorems 7 and 15. $\qquad\square$

**Proposition 21** *The width $w(P)$ of $P$ can be determined in time that is exponential in $w(P)$.*

*Proof.* If $m$ is the number of clauses in $C_I^*(L)$, then there are at most $m^i$ sets of clauses $\mathcal{C}$ in $C_I^*(L)$ such that $|\mathcal{C}| = i$. Each clause in one such set must have at most $n$ literals, where $n$ is the number of fluents in $P$, and hence, if one literal from each clause in $\mathcal{C}$ is collected, we end up with at most $n^i$ sets of literals of size no greater than $i$, some of which are inconsistent with $I$ and some of which are consistent and minimal (no other consistent set in the collection is properly included); both tests being polynomial given that $I$ is in prime implicate form. Thus constructing the cover $c(\mathcal{C})$ for a set of clauses $\mathcal{C}$ with $|\mathcal{C}| = i$ is exponential in $i$, while checking whether one such cover satisfies $C_I(L)$ is a polynomial operation provided that $I$ is in prime implicate form. Indeed, if $c(\mathcal{C}) = \{t_1, \ldots, t_n\}$, computing the closures $t_i^*$ for each $t_i \in c(\mathcal{C})$, when $I$ is in PI, and testing whether each $t_i^*$ intersects each clause in $C_I(L)$ are polynomial operations (the former reducing to checking for each literal $L'$ whether $I \models \neg t_i^* \vee L'$). Thus for computing $width(L)$, we generate all sets $\mathcal{C}$ of clauses in $C_I^*(L)$ with $|\mathcal{C}| = i$, starting with $i = 0$, increasing $i$ one by one until for one such set, $c(\mathcal{C})$ satisfies $C_I(L)$. This computation is exponential in $w(L)$, and the computation over all preconditions and goal literals in $P$ is exponential in $w(P)$. $\qquad\square$

**Proposition 22** *The width of $P$ is such that $0 \leq w(P) \leq n$, where $n$ is the number of fluents whose value in the initial situation is not known.*

*Proof.* The inequality $0 \leq w(P)$ is direct as $w(L)$ is defined as the size $|\mathcal{C}|$ of the minimal set of clauses $\mathcal{C}$ in $C_I^*(L)$ such that $c(\mathcal{C})$ satisfies $C_I(L)$, and $w(P) = w(L)$ for some precondition and goal literal $L$. The inequality $w(P) \leq n$ follows by noticing that for the set $\mathcal{C}$ of clauses given by the tautologies $L' \vee \neg L'$ in $C_I^*(L)$, $c(\mathcal{C})$ must satisfy each clause $c$ in $C_I(L)$, as each $t \in c(\mathcal{C})$ must assign a truth value to each literal in $c$, and if inconsistent with $c$, it will be inconsistent with $I$ and thus pruned from $c(\mathcal{C})$. Finally, the max number of such tautologies in $C_I^*(L)$ is the number of fluents $L'$ such that neither $L'$ nor $\neg L'$ are unit clauses in $I$. $\quad\square$

**Theorem 24** *For a fixed $i$, the translation $K_i(P)$ is sound, polynomial, and if $w(P) \leq i$, covering and complete.*

*Proof.* For soundness, we just need to prove that all merges $m$ in $K_i(P)$ are valid and that all tags $t$ in $K_i(P)$ are consistent. The soundness follows from Theorem 7. The merges $m$ for a literal $L$ in $K_i(P)$ are given by the covers $c(\mathcal{C})$ of collections $\mathcal{C}$ of $i$ or less clauses in $C_i^*(L)$ and clearly since each model $\mathcal{M}$ of $I$ must satisfy $C_i^*(L)$, it must satisfy some $t \in c(\mathcal{C})$ so that $I \models \bigvee_{t \in m} t$ for $m = c(\mathcal{C})$. At the same time, from the definition of the cover $c(\mathcal{C})$, each of these tags $t$ must be consistent with $I$.

For proving that $K_i$ is polynomial for a fixed $i$, we follow ideas similar to the ones used in the proof of Proposition 21 above, where we have shown that the width of $P$ can be determined in time that is exponential in $w(P)$ and polynomial in the number of clauses and fluents in $P$. For a fixed $i$, the number of sets of clauses $\mathcal{C}$ in $C_i^*(L)$ with size $|\mathcal{C}| \leq i$ is polynomial, and the complexity of computing the covers $c(\mathcal{C})$ for such sets, and hence, the merges $m$ for $L$ in $K_i(P)$ is polynomial too. Thus, the whole translation $K_i(P)$ for a fixed $i$ is polynomial in the number of clauses, fluents, and rules in $P$.

Finally, for proving completeness, if $w(P) \leq i$, then $w(L) \leq i$ for each precondition and goal literal $L$ in $P$. Therefore, for each such literal $L$, there is a set $\mathcal{C}$ of clauses in $C_i^*(L)$ such that $c(\mathcal{C})$ satisfies $C_I(L)$. The translation $K_i(P)$ will then generate a unique merge for $L$ that covers $L$. Since $K_i(P)$ is a valid translation, this means that $K_i(P)$ is a covering translation, that is then complete, by virtue of Theorem 15. $\quad\square$

**Lemma A.13.** *If $L'$ is relevant to $L$ and $rel(s, L) \subseteq rel(s', L)$, then $rel(s, L') \subseteq rel(s', L')$.*

*Proof.* If $L''$ is in $rel(s', L')$, then $L''$ is relevant to $L'$, and since $L'$ is relevant to $L$ and the relevance relation is transitive, $L''$ is relevant to $L$. Thus, $L''$ is in $rel(s, L)$ and therefore, since $rel(s, L) \subseteq rel(s', L)$, $L''$ is in $rel(s', L)$. But then $L''$ is in $s'$ and since it is relevant to $L'$, $L''$ is in $rel(s', L')$. $\quad\square$

**Proposition 26** *Let $s$ and $s'$ be two states and let $\pi$ be an action sequence applicable in the classical problems $P/s$ and $P/s'$. Then if $\pi$ achieves a literal $L$ in $P/s'$ and $rel(s', L) \subseteq rel(s, L)$, $\pi$ achieves the literal $L$ in $P/s$.*

*Proof.* By induction on the length of $\pi$. If $\pi$ is empty, and $\pi$ achieves a literal $L$ in $P/s'$, $L$ must be in $s'$, and since $L$ is relevant to itself, $L \in rel(s', L)$. Then as $rel(s', L) \subseteq rel(s, L)$, $L$ must be in $s$, and thus $\pi$ achieves $L$ in $P/s$.

Likewise, if $\pi_{+1} = \pi, a$ achieves $L$ in $P/s'$ then A) there is rule $a : C \rightarrow L$ such that $\pi$ achieves $C$ in $P/s'$; or B) $\pi$ achieves $L$ in $P/s'$ and for any rule $a : C' \rightarrow \neg L$, $\pi$ achieves $\neg L'$ in $P/s'$ for some $L' \in C'$.

If A, $\pi$ must achieve each literal $L_i \in C$ in $P/s'$. Since $L_i$ is relevant to $L$ and $rel(s', L) \subseteq rel(s, L)$, by Lemma A.13, $rel(s', L_i) \subseteq rel(s, L_i)$. Then, by inductive hypothesis, the plan $\pi$ must achieve $L_i$ in $P/s$ for each $L_i \in C$, and thus $\pi_{+1} = \pi, a$ must achieve $L$ in $P/s$

If B, since each such $\neg L'$ is relevant to $L$ (as $L'$ is relevant to $\neg L$), and $rel(s', L) \subseteq rel(s, L)$, by Lemma A.13, $rel(s', \neg L') \subseteq rel(s, \neg L')$, and thus by inductive hypothesis, $\pi$ must achieve $\neg L'$ in $P/s$ and also $L$, so that $\pi_{+1} = \pi, a$ must achieve $L$ in $P/s$. $\qquad \square$

**Lemma A.14.** *If $S$ and $S'$ are two collection of states such that for every state $s$ in $S$ and every precondition and goal literal $L$ in $P$, there is a state $s'$ in $S'$ such that $rel(s', L) \subseteq rel(s, L)$, then if $\pi$ is applicable in $P/S'$, $\pi$ is applicable in $P/S$.*

*Proof.* By induction on the length of $\pi$. If $\pi$ is empty, it is obvious. If $\pi_{+1} = \pi, a$ is applicable in $P/S'$, then $\pi$ is applicable in $P/S'$ and, by inductive hypothesis, $\pi$ is applicable in $P/S$. We need to prove that $\pi$ achieves the preconditions of action $a$ in $P/S$.

For any $L \in Prec(a)$ and any $s \in S$, from the hypothesis, there is a state $s' \in S'$ such that $rel(s', L) \subseteq rel(s, L)$. From Proposition 26, and since $\pi$ achieves $L$ in $P/s'$, $\pi$ must achieve $L$ in $P/s$. Since the argument applies to any $s \in S$, $\pi$ achieves $L$ in $P/S$, and thus $\pi_{+1} = \pi, a$ must be applicable in $P/S$. $\qquad \square$

**Proposition 27** *If $S$ and $S'$ are two collections of states such that for every state $s$ in $S$ and every precondition and goal literal $L$ in $P$, there is a state $s'$ in $S'$ such that $rel(s', L) \subseteq rel(s, L)$, then if $\pi$ is a plan for $P$ that conforms with $S'$, $\pi$ is a plan for $P$ that conforms with $S$.*

*Proof.* From Lemma A.14, we consider a problem $P'$ similar to $P$ but with a new action $a_G$ whose preconditions are the goals $G$ of $P$. If $\pi$ is a plan for $P$ that conforms with $S'$, then the action sequence $\pi, a_G$ is applicable in $P'/S'$, and then from the lemma, $\pi, a_G$ is applicable in $P'/S$, and thus $\pi$ must be a plan for $P/S$ $\qquad \square$

**Proposition 28** *$S'$ is a basis for $P$ if for every possible initial state $s$ of $P$ and every precondition and goal literal $L$ in $P$, $S'$ contains a state $s'$ such that $rel(s', L) \subseteq rel(s, L)$.*

*Proof.* Direct from Proposition 27, by considering $S$ to be the set of possible initial states of $P$. $\qquad \square$

**Proposition 29** *If the initial situation $I$ is in prime implicate form and $m = \{t_1, \ldots, t_n\}$ is a merge that covers a literal $L$ in $P$, then the set $S[t_i, L]$ of possible initial states $s$ of $P$ such that $rel(s, L) \subseteq t_i^*$ is non-empty.*

*Proof.* Direct from Theorem A.10. $\qquad \square$

**Theorem 30** *Let $K_{T,M}(P)$ be a covering translation and let $S'$ stand for the collection of states $s[t_i, L]$ where $L$ is a precondition or goal literal of $P$ and $t_i$ is a tag in a merge $m$ that covers $L$. Then $S'$ is a basis for $P$.*

*Proof.* We show that for every possible initial state $s$ and any precondition and goal literal $L$, $S'$ in the theorem contains a state $s'$ such that $rel(s', L) \subseteq rel(s, L)$. The result then follows from Proposition 28. Indeed, any such state $s$ must satisfy a tag $t_i$ in a covering merge $m = \{t_1, \ldots, t_n\}$ for $L$, as these merges are valid. But from Theorem A.10, there must be a possible initial state $s'$ such that $rel(s', L) \subseteq t_i^*$, and therefore, $rel(s', L) \subseteq rel(s, L)$ as $s$ must satisfy $t_i^*$ and possibly other literals $L'$ that are relevant to $L$. ☐

**Theorem 31** *If $P$ is a conformant planning problem with bounded width, then $P$ admits a basis of polynomial size.*

*Proof.* If $w(P) \leq i$ for a fixed $i$, $K_i(P)$ is a covering translation with a polynomial number of merges and tags, and in such case, the basis $S'$ for $P$ defined by Theorem 30 contains a polynomial number of states, regardless of the number of possible initial states. ☐

## Appendix B. Consistency

We have been assuming throughout the paper that the conformant planning problems $P$ and their translations $K_{T,M}(P)$ are consistent. In this section we make this notion precise, explain why it is needed, and prove that $K_{T,M}(P)$ is consistent if $P$ is. For the proof, we take into account that the heads $KL$ of the merge actions $a_{m,L}$ in $K_{T,M}(P)$, are extended with the literals $K\neg L'$ for the literals $L'$ that are mutex with $L$ in $P$ (see Definition 4).

We start at the beginning assuming that *states are not truth-assignments but sets of literals over the fluents of the language.* A state is *complete* if for every literal $L$, $L$ or $\neg L$ is in $s$, and *consistent* if for no literal both $L$ and $\neg L$ are in $s$. Complete and consistent states represent truth-assignments over the fluents $F$ and the consistency of $P$ and of the translation $K_{T,M}(P)$ ensures that all applicable action sequences $\pi$ map complete and consistent states $s$ into complete and consistent states $s'$. Once this is guaranteed, complete and consistent states can be referred to simply as states which is what we have done in the paper.

Given a complete state $s$ and an action $a$ applicable in $s$, the next state $s_a$ is

$$s_a = (s \setminus Del(a, s)) \cup Add(a, s)$$

where

$$Add(a, s) = \{L \,|\, a : C \rightarrow L \text{ in } P \text{ and } C \subseteq s\}$$

and

$$Del(a, s) = \{\neg L \,|\, L \in Add(a, s)\} \,.$$

It follows from this that $s_a$ is a complete state if $s$ is a complete state, as the action $a$ only 'deletes' a literal $L$ in $s$ if $\neg L$ is added by $a$ in $s$. On the other hand, $s$ may be consistent and $s_a$ inconsistent, as for example, when there are rules $a : C \rightarrow L$ and $a : C' \rightarrow \neg L$ such that both $C$ and $C'$ are in $s$. In order to exclude this possibility, ensuring that all reachable states are complete and consistent, and thus represent genuine *truth assignments* over the fluents in $F$, a consistency condition on $P$ is needed:

**Definition B.1** (Consistency). *A classical or conformant problem $P = \langle F, I, O, G \rangle$ is consistent if the initial situation $I$ is logically consistent and every pair of complementary literals $L$ and $\neg L$ are mutex in $P$.*

In a *consistent* classical problem $P$, all the reachable states are complete and consistent, and the standard progression lemma used in the preceding proofs holds:

**Theorem B.2** (Progression). *An action sequence $\pi_{+1} = \pi, a$ applicable in the complete and consistent state $s$ achieves a literal $L$ in a* consistent *classical problem $P$ iff A) $\pi$ achieves the body $C$ of a rule $a : C \to L$ in $P$, or B) $\pi$ achieves $L$ and for every rule $a : C' \to \neg L$, $\pi$ achieves $\neg L'$ for a literal $L'$ in $C'$.*

We will see below that if a conformant problem $P$ is consistent in this sense, so will be any valid translation $K_{T,M}(P)$. We have tested all the benchmarks considered in this paper for consistency and found all of them to be consistent except for two domains that we have introduced elsewhere: 1-Dispose and Look-and-Grab. In these cases, since the consistency of the classical problem $K_{T,M}(P)$ cannot be inferred from the consistency of $P$, it can be checked explicitly using Definition B.1, or similarly, the plans that are obtained from $K_{T,M}(P)$ can be checked for consistency as indicated in Section 8: the soundness of these plans is ensured provided that they never trigger conflicting effects $KL/t$ and $\neg KL/t$.[10]

*Proof.* The proof of Theorem B.2 does not rest on a particular definition of mutexes, just that mutex atoms are not both true in a reachable state. In a consistent problem $P$, an applicable action sequence $\pi$ maps $s$ into a complete and consistent state $s'$ that represents a truth assignment. Then, the action sequence $\pi_{+1} = \pi, a$ achieves $L$ iff C) $L \in Add(a, s')$ or D) $L \in s'$ and $\neg L \notin Del(a, s')$. Condition A in the theorem, however, is equivalent to C, and Condition B in the theorem, is equivalent to D. Indeed, $L \notin Del(a, s')$ iff for each rule $a : C' \to \neg L$ there is a literal $L' \in C'$ such that $L' \notin s'$, which, given that $s'$ is complete and consistent, is true iff $\neg L' \in s'$ (this is precisely where consistency is needed; else $\neg L' \in s'$ would not imply $L' \notin s'$). $\qquad\square$

The notion of mutex used in the definition of consistency expresses a guarantee that a pair of literals is not true in a reachable state. Sufficient and polynomial conditions for mutual exclusivity and other type of invariants have been defined in various papers, here we follow the definition by Bonet and Geffner (1999).

**Definition B.3** (Mutex Set). A mutex set is a collection $R$ of unordered literals pairs $(L, L')$ over a *classical* or *conformant* problem $P$ such that:

1. for no pair $(L, L')$ in $R$, both $L$ and $L'$ are in a possible initial state $s$,

2. if $a : C \to L$ and $a : C' \to L'$ are two rules for the same action where $(L, L')$ is a pair in $R$, then $Pre(a) \cup C \cup C'$ is mutex in $R$, and

3. if $a : C \to L$ is a rule in $P$ for a literal $L$ in a pair $(L, L')$ in $R$, then either a) $L' = \neg L$, b) $Pre(a) \cup C$ is mutex with $L'$ in $R$, or c) $Pre(a) \cup C$ implies $C'$ in $R$ for a rule $a : C' \to \neg L'$ in $P$;

---

10. The consistency of the two domains, 1-Dispose and Look-and-Grab, can be established however if a definition of mutexes slightly stronger than the one below is used. It actually suffices to change the expression $Pre(a) \cup C$ in clause 3c) of the definition of mutex sets below by $Pre(a) \cup C \cup \{L'\}$.

In this definition, a pair is said to be mutex in $R$ if it belongs to $R$, a set of literals $S$ is said to be mutex in $R$ if $S$ contains a pair in $R$, and a set of literals $S$ is said to imply a set of literals $S'$ in $R$ when $S$ is mutex in $R$ with the complement $\neg L$ of each literal $L$ in $S' \setminus S$.

It easy to verify that if $R_1$ and $R_2$ are mutex sets, their union $R_1 \cup R_2$ is a mutex set, and thus that there is a maximal mutex set for $P$ that we denote as $R^*$. The pairs in $R^*$ are just called *mutexes*.

For simplicity and without loss of generality, we will assume that preconditions $Pre(a)$ are empty. Indeed, it is simple to show that the mutexes of a problem $P$ remain the same if preconditions are pushed in as conditions. We also assume that no condition $C$ in a rule $C \to L$ in $P$ is mutex, as these rules can be simply pruned. In addition, we assume that no literal $L$ is mutex with a pair of complementary literals $L'$ and $\neg L'$, as then $L$ cannot be true in a reachable state, and thus, can be pruned as well.

The definition of mutexes is sound, meaning that no pair in a mutex set can be true in a reachable state:

**Theorem B.4.** *If $(L, L')$ is a pair in a mutex set $R$ of a classical or conformant problem $P$, then for no reachable state $s$ in $P$, $\{L, L'\} \subseteq s$.*

*Proof.* We proceed inductively. Clearly, $L$ and $L'$ cannot be part of a possible initial state, as this is ruled out by the definition of mutex sets. Thus, let us assume as inductive hypothesis that $L$ and $L'$ are not part of any state $s$ reachable in less than $i$ steps, and let us prove that the same is true for the states $s' = s_a$ that are reachable from $s$ in one step. Clearly if $L$ and $L'$ belong to $s'$, then either A) both $L$ and $L'$ belong to $Add(a, s)$, or B) $L$ belongs to $Add(a, s)$ and $L'$ belongs to $s$ but not to $Del(a, s)$. We show that this is not possible. For A, $P$ must comprise rules $a : C \to L$ and $a : C' \to L'$ such that $C \cup C' \subseteq s$, yet from the definition of mutex sets, $C \cup C'$ must be mutex, and from the inductive hypothesis then $C \cup C' \not\subseteq s$. For B, there must be a rule $a : C \to L$ with $C \subseteq s$, but then from $L' \in s$ and the inductive hypothesis, it follows that $L'$ is not mutex with $C$ in $R$, and thus, from the mutex set definition, that either $L' = \neg L$ or $C$ implies $C'$ for a rule $a : C' \to \neg L'$. In the first case, however, due to the rule $a : C \to L$ and $C \subseteq s$, $L' \in Del(a, s)$, while in the second case, from the completeness of all reachable states, we must have $C' \subseteq s$, and hence $L' \in Del(a, s)$, contradicting B in both cases. $\square$

Provided that the initial situation $I$ of a conformant planning problem $P$ is in prime implicate form, computing the largest mutex set $R^*$ and testing the consistency of $P$ are polynomial time operations. For the former, one starts with the set of literal pairs and then iteratively drops from this set the pairs that do not comply with the definition until reaching a fixed point (Bonet & Geffner, 1999).

We move on now to prove that if a conformant problem $P$ is consistent, so is a valid translation $K_{T,M}(P)$. The consistency of the classical problems $P/s$ for possible initial states $s$ is direct, as the set of mutexes in $P$ is a subset of the set of mutexes in $P/s$ where the initial situation is more constrained.

**Proposition B.5** (Mutex Set $R_T$). *For a valid translation $K_{T,M}(P)$ of a consistent conformant problem $P$, define $R_T$ to be the set of (unordered) literals pairs $(KL/t, KL'/t')$ and $(KL/t, \neg K \neg L'/t)$ where $(L, L')$ is a mutex in $P$, and $t$ and $t'$ are two tags jointly satisfiable with $I$ ($I \not\models \neg(t \cup t')$). Then $R_T$ is a mutex set in $K_{T,M}(P)$.*

It follows from this that $K_{T,M}(P)$ is consistent if $P$ is consistent, as then $L' = \neg L$ is mutex with $L$ in $P$, and so $(KL/t, \neg KL/t)$ must be a mutex in $R_T$.

**Theorem B.6** (Consistency $K_{T,M}(P)$). *A valid translation $K_{T,M}(P)$ is consistent if $P$ is consistent.*

The consistency of the translation $K_0(P)$ follows as a special case, as $K_0(P)$ is $K_{T,M}(P)$ with an empty set of merges $M$ and a set of tags $T$ containing only the empty tag. We are left to prove Proposition B.5.

*Proof of Proposition B.5.* We must show that the set $R_T$ comprised of the pairs $(KL/t, KL'/t')$ and $(KL/t, \neg K \neg L'/t)$ for $L'$ mutex with $L$ in $P$, and tags $t$ and $t'$ jointly satisfiable with $I$, is a set that complies with clauses 1, 2, and 3 of Definition B.3. We go one clause at a time.

1. No pair in $R_T$ can be true initially in $K_{T,M}(P) = \langle F', I', O', G' \rangle$ for jointly satisfiable $I$, $t$, and $t'$. Indeed, if both $KL/t$ and $KL'/t'$ are in $I'$ there must be a possible initial state satisfying $t$ and $t'$ where $L$ and $L'$ are true in contradiction with $L$ and $L'$ being mutex in $P$. Similarly, if $KL/t$ is in $I'$ but $K \neg L'/t$ not, it must be the case that $I \models t \supset L$ and $I \not\models t \supset \neg L'$, so that there must be some possible initial state of $P$ where $t$, $L$, and $L'$ hold, a contradiction with $L$ and $L'$ being mutex in $P$ too.

2. If there is an action $a$ with rules for $KL/t$ and $KL'/t'$ then the rules must be support rules of the form $a : KC/t \rightarrow KL/t$ and $a : KC'/t' \rightarrow KL'/t'$ arising from rules $a : C \rightarrow L$ and $a : C' \rightarrow L'$ in $P$.[11] Then since $L$ and $L'$ are mutex in $P$, $C$ and $C'$ must contain literals $L_1 \in C$ and $L_2 \in C'$ such that $(L_1, L_2)$ is a mutex in $P$, and hence $(KL_1/t, KL_2/t')$ belongs to $R_T$, so that $KC/t$ and $KC'/t'$ are mutex in $R_T$ as well.

   Similarly, if there is an action with rules for $KL/t$ and $\neg K \neg L'/t$ for a literal $L'$ mutex with $L$ in $P$, the rules must be support and cancellation rules of the form $a : KC/t \rightarrow KL/t$ $a : \neg K \neg C'/t \rightarrow \neg K \neg L'/t$, arising from rules $a : C \rightarrow L$ and $a : C' \rightarrow L'$ in $P$. Since $L$ and $L'$ are mutex in $P$, $C$ and $C'$ must contain literals $L_1 \in C$ and $L_2 \in C'$ that are mutex in $P$, and hence $R_T$ must contain the pair $(KL_1/t, \neg K \neg L_2/t)$, so that $KC/t$ and $\neg K \neg C'/t$ must be mutex in $R_T$.

3. We are left to show that the set $R_T$ given by the pairs $(KL/t, KL'/t')$ and $(KL/t, \neg K \neg L'/t)$ complies with clause 3 in the definition of mutex sets as well. Consider the first class of pairs $(KL/t, KL'/t')$ and a rule $a : KC/t \rightarrow KL/t$ for $KL/t$ arising from a rule $a : C \rightarrow L$ in $P$. Since $L$ is mutex with $L'$ in $P$, then one of the conditions 3a, 3b, or 3c must hold for the rule $a : C \rightarrow L$ and $L'$. If 3a, then $L' = \neg L$, and $KC/t$ must imply the body $\neg K \neg C/t'$ of the cancellation rule $a : \neg K \neg C/t' \rightarrow \neg K \neg L/t'$, as for each literal $L_1$ in $C$, $R_T$ must contain the pair $(KL_1/t, K \neg L_1/t')$ so that $KL_1/t$ implies $\neg K \neg L_1/t'$, and $KC/t$ implies $\neg K \neg C/t'$ (case 3c). If 3b, then $C$ and $L'$ are

---

11. The action $a$ cannot be a merge for a literal $L''$ mutex with both $\neg L$ and $\neg L'$, as in such case, $L''$ implies that $L$ and $L'$ that are mutex. Similarly, $a$ cannot be a merge for $L$ as in such a case, $L$ will be mutex with both $L'$ and $\neg L'$. For the same reason, $a$ cannot be a merge for $L'$ either. Thus, the action $a$ above cannot be a merge and must be an action from $P$.

mutex in $P$, and thus $C$ contains a literal $L_1$ mutex with $L'$ in $P$. This means that the pair $(KL_1/t, KL'/t')$ is in $R_T$ and hence that $KC/t$ is mutex with $KL'/t'$ in $R_T$ (case 3b). Last, if 3c, $C$ implies $C'$ in $P$ for a rule $a : C' \rightarrow \neg L'$, but then $KC/t$ must imply the body $\neg K \neg C'/t'$ of the cancellation rule $a : \neg K \neg C'/t' \rightarrow \neg KL'/t'$. Indeed, for each literal $L_1$ in both $C$ and $C'$, we had above that $KL_1/t$ implies $\neg K \neg L_1/t'$, while if $L_2$ is a literal in $C'$ but not in $C$, then some literal $L_3 \in C$ must be mutex with $\neg L_2$ in $P$, and hence the pair $(KL_3/t, K \neg L_2/t')$ must be in $R_T$ and $KL_3/t$ implies then $\neg K \neg L_2/t'$ (case 3c)

Consider now the same pair $(KL/t, KL'/t')$ along with a merge action $a_{m,L}$ with a rule $\bigwedge_{t_i \in m} KL/t_i \rightarrow KL$ for $KL/t = KL$ (thus $t$ is the empty tag). In this case, since the merge $m$ is valid and $t'$ is consistent, there must be some $t_i \in m$ such that $t_i$ and $t'$ are jointly consistent with $I$. It follows then that $(KL/t_i, KL'/t')$ is a pair in $R_T$ and thus that the body of the merge is mutex with $KL'/t'$ in $R_T$ (case 3b).

There is no need to consider the pair $(KL/t, KL'/t')$ along with the rules for $KL'/t'$, as the literals $KL/t$ and $KL'/t'$ have the same structure, and thus the same argument above applies, replacing $t$ with $t'$ and $L$ with $L'$.

We switch now to the second class of pairs $(KL/t, \neg K/\neg L'/t)$ and the rules $a : KC/t \rightarrow KL/t$ for $KL/t$. Since $L$ and $L'$ are mutex in $P$, then conditions 3a, 3b, or 3c must hold. If a, then $L' = \neg L$, and in such a case, condition 3c holds in $K_{T,M}(P)$ as $KC/t$ implies the body $KC/t$ of the rule $a : KC/t \rightarrow K \neg L'$ $(\neg L' = L)$. If b, $C$ is mutex with $L'$, and thus there is a literal $L_1$ in $C$ such that $L_1$ and $L'$ are mutex in $P$, and therefore $KC/t$ and $KL'/t$ are mutex in $R_T$ (case 3b). Finally, if c, $C$ implies $C'$ for a rule $a : C' \rightarrow \neg L'$ in $P$, then $KC/t$ must imply $KC'/t$ in $R_T$ for a rule $a : KC'/t \rightarrow K \neg L'/t$ (case 3c).

For the empty tag $t$, the rule for $KL/t$ may also be a merge, but then due to the extra effects $K \neg L'$ in the merge action for $L$, the merge for $KL$ is also a merge for $K \neg L'$, and then case 3c holds.

Last, for the same class of pairs, the only rules for $\neg K \neg L'/t$ are cancellation rules of the form $a : \neg K \neg C''/t \rightarrow \neg K \neg L'/t$ for a rule $a : C'' \rightarrow L'$ in $P$. Since $L'$ is mutex with $L$ in $P$, then conditions 3a, 3b, or 3c must hold for the rule $a : C'' \rightarrow L'$ and $L'$ in $P$. If a, then $L = \neg L'$, and the cancellation rule is then $a : \neg K \neg C''/t \rightarrow \neg KL$ (case 3c). If b, $C''$ is mutex with $L$, and thus there is a literal $L_2$ in $C''$ such that $(L_2, L)$ is a mutex in $P$, and therefore $KL/t$ implies $K \neg L_2/t$ in $R_T$, and hence $\neg K \neg L_2/t$ and $\neg K \neg C''/t$ imply $\neg KL/t$ in $R_T$ (case 3b). Finally, if c, $C''$ implies $C'$ for a rule $a : C' \rightarrow \neg L$ in $P$, and then $\neg K \neg C''/t$ must imply $\neg K \neg C'/t$ for a rule $a : \neg K \neg C'/t \rightarrow \neg KL/t$ in $R_T$. Indeed, if $L_A$ implies $L_B$ in $P$, $\neg L_B$ implies $\neg L_A$ in $P$, and $K \neg L_B/t$ implies $K \neg L_A/t$ in $R_T$, and $\neg K \neg L_A/t$ implies $\neg K \neg L_B/t$.

$\square$